%% file: 0_main.tex
\documentclass{article}

\usepackage{microtype}
\usepackage{graphicx}
\usepackage{subcaption}
\usepackage{booktabs} 

\usepackage{algorithm}
\usepackage{algorithmic}

\usepackage{hyperref}

\usepackage[accepted]{icml2024}

\usepackage{amsmath}
\usepackage{amssymb}
\usepackage{mathtools}
\usepackage{amsthm}

\usepackage[capitalize,noabbrev]{cleveref}
\theoremstyle{plain}
\newtheorem{theorem}{Theorem}[section]

\theoremstyle{definition}

\theoremstyle{remark}

\usepackage[textsize=tiny]{todonotes}

\usepackage{url}            
\usepackage{amsfonts}       
\usepackage{xcolor}         
\definecolor{Red}{rgb}{0.768, 0.054, 0.054}
\definecolor{Blue}{rgb}{0.152, 0.294, 0.925}
\definecolor{Green}{rgb}{0,0.4,0.7}
\definecolor{figred}{RGB}{255, 181, 164}

\usepackage{colortbl}

\usepackage{tikz}
\usepackage[most]{tcolorbox}
\usetikzlibrary{fit,calc}
\newcommand{\boxit}[2]{
    \tikz[remember picture,overlay] \node (A) {};\ignorespaces
    \tikz[remember picture,overlay]{\node[yshift=3pt,fill=#1,opacity=.25,fit={($(A)+(0,0.15\baselineskip)$)($(A)+(.9\linewidth,-{#2}\baselineskip - 0.25\baselineskip)$)}] {};}\ignorespaces
}

\definecolor{yellow}{rgb}{1, 0.8, 0.2}
\definecolor{green}{rgb}{0.5, 0.8, 0.5}
\definecolor{blue}{rgb}{0.4, 0.7, 1}
\usepackage{arydshln}

\usepackage{bbm}
\usepackage{adjustbox}
\usepackage{bm}
\usepackage{caption}
\usepackage{comment}
\usepackage{subcaption}
\usepackage{float}
\usepackage{makecell}
\usepackage{threeparttable}
\usepackage{multirow}
\usepackage{wrapfig}
\usepackage{tabularx}
\usepackage{paralist}
\input{math_commands}
\usepackage{siunitx}
\icmltitlerunning{Mixture-of-Prompts}

\begin{document}

\twocolumn[
\icmltitle{One Prompt is not Enough:\\ Automated Construction of a Mixture-of-Expert Prompts}



\icmlsetsymbol{equal}{*}

\begin{icmlauthorlist}
\icmlauthor{Ruochen Wang}{equal,ucla}
\icmlauthor{Sohyun An}{equal,kaist}
\icmlauthor{Minhao Cheng}{psu}
\icmlauthor{Tianyi Zhou}{umd}
\icmlauthor{Sung Ju Hwang}{kaist}
\icmlauthor{Cho-Jui Hsieh}{ucla} \\
\url{https://github.com/turningpoint-ai/mixture-of-prompts}\\
\end{icmlauthorlist}

\icmlaffiliation{ucla}{University of California, Los Angeles}
\icmlaffiliation{psu}{Penn State University}
\icmlaffiliation{umd}{University of Maryland, College Park}
\icmlaffiliation{kaist}{Korea Advanced Institute of Science \& Technology}

\icmlcorrespondingauthor{Cho-Jui Hsieh}{chohsieh@cs.ucla.edu}

\icmlkeywords{Machine Learning, ICML}

\vskip 0.3in

\begin{center}
\vspace{-16mm}
\centerline{\includegraphics[width=0.5\linewidth]{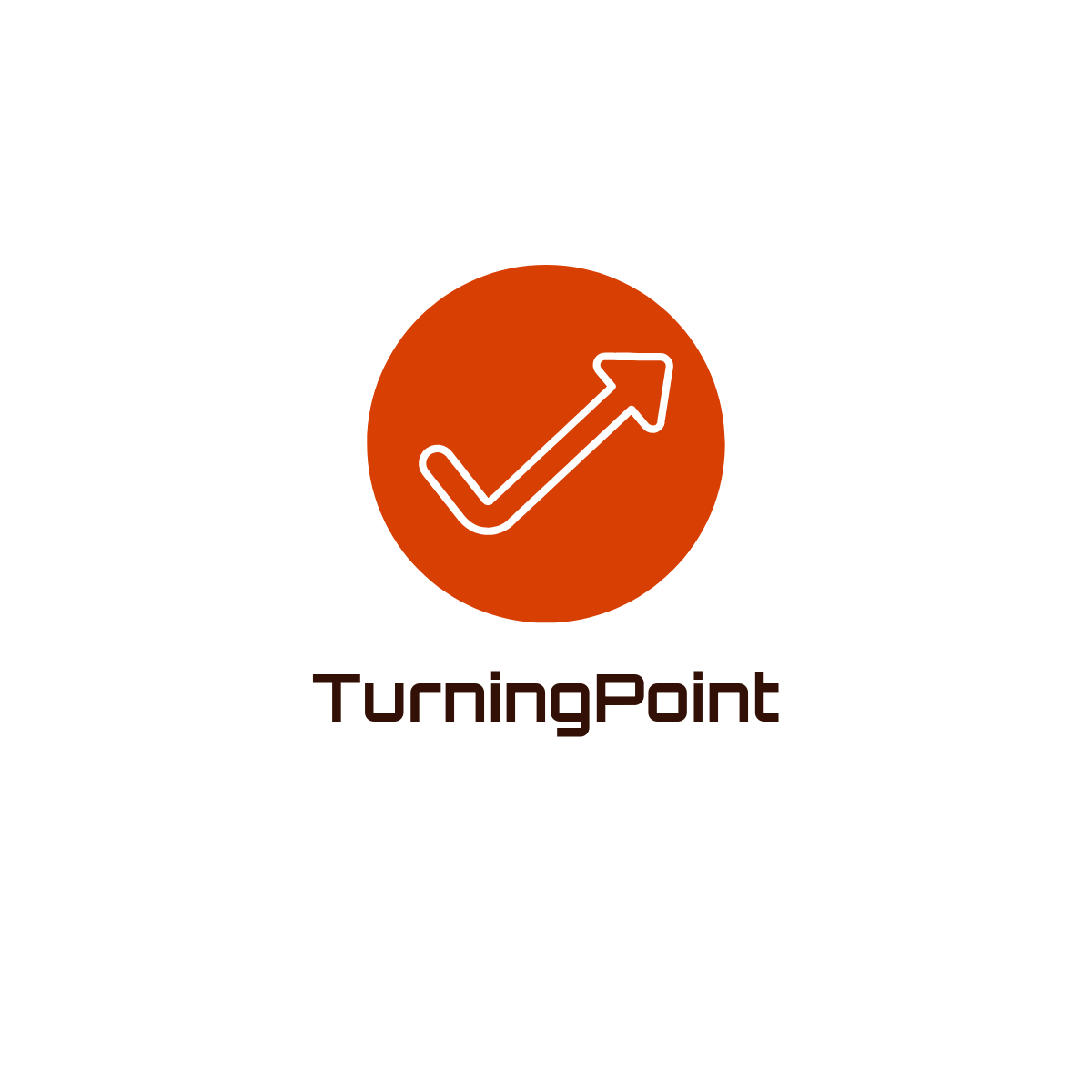}}
\vspace{-19mm}
\end{center}
]



\printAffiliationsAndNotice{\icmlEqualContribution} 

\input{1_abstract}
\input{2_introduction}
\input{3_related_work}
\input{4_method}
\input{5_experiment}
\input{6_ablate}
\input{7_conclusion}

\bibliography{ref}
\bibliographystyle{icml2024}

\input{appendix}

\end{document}

%% file: math_commands.tex
\DeclareMathOperator*{\argmin}{argmin}
\DeclareMathOperator*{\argmax}{argmax}

\renewcommand{\algorithmiccomment}[1]{\hfill{ $\rhd$ #1}}

\newenvironment{sizeddisplay}[1]
 {\par\nopagebreak#1\noindent\ignorespaces}
 {\nopagebreak\ignorespacesafterend}


%% file: 1_abstract.tex

\begin{abstract}
Large Language Models (LLMs) exhibit strong generalization capabilities to novel tasks when prompted with language instructions and in-context demos. Since this ability sensitively depends on the quality of prompts, various methods have been explored to automate the instruction design. While these methods demonstrated promising results, they also restricted the searched prompt to one instruction. Such simplification significantly limits their capacity, as a single demo-free instruction might not be able to cover the entire complex problem space of the targeted task. To alleviate this issue, we adopt the Mixture-of-Expert paradigm and divide the problem space into a set of sub-regions; Each sub-region is governed by a specialized expert, equipped with both an instruction and a set of demos. A two-phase process is developed to construct the specialized expert for each region: (1) \textit{demo assignment}: Inspired by the theoretical connection between in-context learning and kernel regression, we group demos into experts based on their semantic similarity; (2) \textit{instruction assignment}: A region-based joint search of an instruction per expert complements the demos assigned to it, yielding a synergistic effect. The resulting method, codenamed Mixture-of-Prompts (MoP), achieves an average win rate of 81\% against prior arts across several major benchmarks.
\end{abstract}

%% file: 2_introduction.tex
\section{Introduction}
\label{sec:intro}

Recent advancements in large language models (LLMs) have demonstrated a remarkable ability to solve novel tasks described by user instructions~\citep{chatgpt, gpt4, llama, peters2018dissecting, devlin2018bert, brown2020language, wei2022chain}.
Despite the success, there still exists a substantial gap between user intention and the model's interpretation.
Therefore, carefully designed prompts (a.k.a. Prompt Engineering) become an essential ingredient for fully eliciting LLM's superior generalization ability~\citep{alhoshan2022zero, zhao2021calibrate, liu2021makes, lu2021fantastically, su2022selective, wang2022self, cot, tot, tooluse, stepbystep}.
However, it usually requires laborious efforts through inefficient trial and error.
To reduce human efforts, several recent attempts have shown tremendous potential in utilizing LLMs themselves to design prompts for language generation~\citep{zhou2022large, apo, chen2023instructzero, promptbreeder, opro, gps}.
These methods are part of a broader conceptual framework termed \textbf{``LLM as Optimizers''} ~\citep{opro}.
While the results are promising, pioneering efforts along this line primarily focus on finding the optimal demo-free instruction for a specified task, based on a set of (input, output) demonstrations.
While the prompts produced by these methods can outperform human-designed counterparts, a single demo-free instruction might not suffice to serve all the possible instances of a task or cover the whole problem space, limiting the LLM's problem-solving potential.

This paper aims to expand the problem space coverage for automatic prompting by optimizing a Mixture of Prompts (MoP).
Our key insight is to adopt the \textbf{Mixture of Experts (MoE)} paradigm~\citep{moe1, moe2} to partition the problem space into multiple homogeneous regions, each governed by \textbf{a specialized expert (prompt)}.
At inference time, a single expert will be selected to prompt the LLM to answer a new input query.
Under the MoE framework, prompt optimization reduces to an \textbf{expert assignment problem} that aims to search for the most suitable prompt for each expert, with the goal of \textbf{optimizing the performance of their mixture as a whole}.

Another primary improvement proposed in this paper is to expand the prompt of each expert to contain both the instruction and demos, jointly optimized for each expert region in the problem space.
Intuitively, concrete demos are good at providing fine-grained knowledge and expertise (local information) matching the details of input queries in a local region, whereas the instruction provides a generic ability and high-level guidance to solve a task (global information);
Hence, they are complementary and together empower the experts to excel at their problem region.
Motivated by this, we adopt a two-phase search algorithm that jointly optimizes a (demos, instruction) pair per expert: 
We first cluster all demos into different experts according to their semantic similarity, and then search for the best instruction complementary to each demo cluster of a prompt.
For the first phase, i.e., demo assignment, we cluster the demos to multiple regions in a semantic embedding space by clustering algorithms.
For the second phase, i.e., instruction assignment, we introduce a region-based joint search that finds the best instruction to complement the demos assigned to each expert.
Given a new test query, we routine the expert containing the semantically closest demos to it.
This method is inspired by the recently established theoretical connection between In-Context Learning and Kernel Regression~\citep{han2023context}, which suggests that demos semantically closer to a test input in the LLM's embedding space tends to perform better at inferring its answer.

We scrutinize the proposed Mixture-of-Prompts (MoP) through extensive empirical study.
Our key findings can be summarized as follows:
(1) Clustering demos in the embedding space can effectively find semantically similar clusters that help allocate test samples accurately to the corresponding region and the optimal expert. 
(2) More experts are not necessarily better: there exists an optimal number of partitions for the problem space.
(3) The optimal instruction for each demo cluster is often distinct, necessitating the joint search of demo and instructions.
We further validate the strength of MoP across three major prompt optimization benchmarks: Instruction-Induction~\cite{ii}, Super Natural Instructions~\cite{superni}, and BIG-Bench-Hard~\cite{bbh}.
These benchmarks cover a wide range of possible tasks, including coding, math, common-sense reasoning, knowledge retrieval, etc.
The results show that MoP surpasses six representative recent methods, achieving an average win rate of 81\% across several major benchmarks.
Our key contribution can be summarized as follows:
\begin{compactitem}
    \item We propose a Mixture-of-Prompt (MoP), a Mixture-of-Expert framework that partitions the problem space into homogenous regions.
    \item We extend each expert prompt to contain both instruction and demos, which expand the output space of prompt optimization. 
    \item Our empirical study with 50 tasks - one of the largest in prompt optimization literature - reveals that the proposed two-step search algorithm, which leverages semantic similarity for demo assignment and region-based joint search for instruction assignment, achieves significant performance gains on major benchmarks.
\end{compactitem}

%% file: 3_related_work.tex
\section{Related work}
\label{sec:related_work}

\paragraph{Prompt optimization for language generation.}
Aligning pretrained language models with human intentions is a crucial step toward unlocking their potential~\citep{chatgpt, tooluse, stepbystep}.
An effective line of training-free alignment methods is prompt optimization (PO)~\citep{autoprompt, zhou2022large}.
PO originated from in-context learning (ICL)~\citep{gpt3}, which is mainly concerned with various designs and arrangements of in-context demonstrations~\citep{cot, tot}.
It later evolves into automatic prompt engineering, where various discrete optimization algorithms are utilized to search for the best prompt~\citep{autoprompt, rlprompt, tempera}.
With the emergence large language models (LLMs), there has been a paradigm shift towards leveraging these models for optimizing prompts in a manner akin to human writers~\citep{zhou2022large, apo, gps, opro, chen2023instructzero, promptbreeder}.
Our research builds on this recent advancement as these method yields strong results and offers a more interpretable optimization process.

\paragraph{Mixture of Experts Paradigm.}
Mixture of Experts~\citep{moe1, moe2} is a classic paradigm of longstanding interest within the machine learning community.
MoE structure was originally studied based on traditional machine learning models~\citep{moe-old1, moe-old2}.
Subsequently, it was extended to deep neural networks by \citep{moe-nn-base} to enhance its capacity to handle complex vision and speech problems.
Following this development, there has been a proliferation of MoE layers integrated with various base neural network structures~\citep{moe-nn1,moe-nn2,moe-nn3}, leading to significant accomplishments in a wide range of language-related tasks.
In recent years, efforts to combine the MoE layer with various base network architectures have demonstrated remarkable successes in modeling natural languages.
Our work extends this high-level paradigm developed in the architectural domain to the prompt optimization task, where each expert is defined as a specialized prompt.

%% file: 4_method.tex
\section{Preliminaries}
\label{sec:prelim}
\input{figures_tex/mop}

\paragraph{Terminology.}
We start by introducing key terminologies that will be used throughout the paper.
We define a \textbf{\color{red}Prompt} as the entire text preceding the question.
We consider the setting where a prompt can be divided into two parts: (1) \textbf{\color{cyan}Instruction}: a set of natural language sentences describing the task, and (2) \textbf{\color{orange}Demos}: a set of input-output pairs structured in a specific way to demonstrate how to solve a task.
Below is an example prompt under this definition:
\begin{align}
    \text{\textit{\color{red}Prompt =}} & \text{\textit{\color{cyan}"Find the opposite words of the input.}} \notag \\
    \text{} & \text{\textit{\color{orange}Input: Similar Output: Dissimilar ..."}} \notag
\end{align}
Mathematically, a prompt ($P$) can be represented as follows~\citep{xie2021explanation}:
\begin{equation}
\begin{aligned}
    P(x) &= \left[I, x_1,\ y_1,\ o^{\text{delim}}, ...,\ x_n,\ y_n,\ o^{\text{delim}},\ x\right]. 
\end{aligned} \label{eq:prompt_def}
\end{equation}
Here, $I$ represents an instruction, ${\{(x_i, y_i)\}}_{i=1}^{n}$ represents in-context demos, which is the set of (input, output) pairs, and $o^{\text{delim}}$ represents delimiter token. 

\paragraph{Prompt Optimization.}
Recent efforts have demonstrated significant potential in automating the prompt engineering processes.
Concretely, given a set of demos sampled from a task distribution $\mathcal{D}$, analog to the ``training data" in supervised learning, a prompt optimization aims at finding an Instruction (demo-free) that minimizes the empirical risk (or maximizes a score):
\begin{align}
    & P^*(x) = \argmax_{P(x)} {\mathbb{E}_{(x, y) \sim \mathcal{D}}{f\left(P(x), y\right)}}, 
\label{eq:auto_prompting_goal}
\end{align}
where $f(\cdot)$ denotes some task-specific scoring function (\Cref{appendix:score_functions}).
After optimization, the best instruction can be used to predict new inputs in the following format: $P^*(x) = \left[I^*,\ x\right]$.
\textbf{Under the framework of Empirical Risk Minimization, one can deduce an underlying assumption that demos (training data) 
encapsulate all external information about the task.}

\paragraph{APE - Automatic Prompt Engineering.}
The most relevant work to ours is APE~\citep{zhou2022large} - a pioneering method demonstrating that LLMs can be used to optimize prompt.
The key idea of APE is to ask an LLM to induce candidate instructions by observing a subset of demos, randomly sampled from the entire training dataset, and pick the best one according to their rank on a held-out validation set (partitioned from training demos as well). Formally,
\begin{align}
    & \left\{I^j\right\}_{j=1}^m \sim \mathbb{P}\left(I^j \mid \Tilde{\mathcal{D}}^{\text{train}}, T(\Tilde{\mathcal{D}}^{\text{train}}); \mathcal{M}_{\phi} \right).
\label{eq:gen_prompts}
\end{align} 
Here, $\mathcal{M}_{\phi}$ denote an LLM, $\{I^j\}_{j=1}^{m}$ are the candidate instructions, and $T(\Tilde{\mathcal{D}}^{\text{train}})$ represents the chosen template format (see \Cref{fig:template}).
Subsequently, the best instruction among the candidate pool is selected based on the validation accuracy:
\begin{align}
    & I^* = \argmax_{I^j} {\mathbb{E}_{(x, y) \sim \Tilde{\mathcal{D}}^{\text{valid}}}{f\left(\left[I^j,\ x\right], y\right)}}.
\label{eq:val_prompts}
\end{align} 
We refer the reader to \Cref{appendix:sub:backgrouds_ape} for more details.

\paragraph{Limitations of APE.}
While methods like APE demonstrated promising results in designing prompts that surpass human engineers, they are still constrained to searching within a \emph{single demo-free} instruction space.
Such a limitation can hinder the problem-solving potential in NLP tasks, where the complexity of problems may not be adequately addressed by a \emph{single demo-free} instruction alone.


\section{\textbf{MoP}: Mixture-of-Prompts}
\label{sec:method}

\subsection{Framework Overview}
\label{sec:method-overview}
\paragraph{Mixture-of-Expert for prompt optimization.}
To address the aforementioned issue of existing automatic prompt engineering methods, we expand the problem space coverage for automatic prompt engineering by optimizing the Mixture of Prompts (MoP). To achieve this, we employ the Mixture of Experts (MoE) paradigm~\citep{moe1, moe2} to partition the entire problem space into $C$ regions, each governed by a specialized expert. Within the MoE framework, prompt optimization (\Cref{eq:auto_prompting_goal}) transforms into an \textbf{expert assignment problem} that aims to search for the most suitable prompt $P_{c}^*$ for each expert, with the ultimate goal of optimizing the performance of the entire mixture:
\begin{equation}
\begin{aligned}
    P^*(x) &= \argmax_{\{P_1(x), \ldots, P_C(x)\}} {\sum_{c=1}^{C} {\mathbb{E}_{(x_c, y_c) \sim \mathcal{V}_{c}}{f\left(P_c(x_c), y_c\right)}}}, \\
    & \quad \text{ where } \mathcal{D} = \{\mathcal{V}_{c}\}_{c=1}^{C}.
\end{aligned} \label{eq:auto_prompting_goal_mop}
\end{equation}
Here, $(x_c, y_c) \sim \mathcal{V}_c$ refers to the data point assigned to expert $c$ by the employed routing function during inference time (we explain it in more detail later in \Cref{sec:method-demo_assign}).
Notably, our MoP framework expands the prompt for each expert to contain both the instruction and demos jointly optimized for each expert region in the problem space; $P_c(x_c) = \left[I_c,\ {{\mathcal{V}}_{c}^{\text{train}}},\ x_c\right]$ in \Cref{eq:auto_prompting_goal_mop}. Intuitively, concrete demos excel at defining fine-grained details and expertise (local information) matching the queries in a local region, whereas instructions provide general abilities and high-level explanations for solving tasks (global information). Inspired by this, we introduce a \textbf{two-phase search algorithm} that jointly optimizes (demos, instructions) pairs for each expert (detail in \Cref{sec:method-demo_assign} and \ref{sec:method-inst_assign}).

\subsection{Demo Assignment}
\label{sec:method-demo_assign}
In our two-phase search algorithm, we initiate the process by assigning training demos to different experts.
Since demos represent local expertise, their assignment defines the design of experts in MoE and is entirely up to the constructors.
While there are many options, we propose a clustering-based demo assignment method, derived from a recent theory of In-Context Learning in LLMs.

\paragraph{LLM learns from demos via Kernel Regression in the embedding space.}
Recently, ~\citet{han2023context} provides the first theoretical result showing that LLM performs In-Context Learning (ICL) from demos as Kernel Regression in the embedding space.
Formally:
\begin{theorem}
\label{theorem:icl}
    Let $\{(x_i, y_i)\}_{i=1}^n$ denote the demos used in the prompt; Let $K$ define a kernel function that measures the semantic similarity between two data points, which can be represented as $K(x_i, x_j)= \phi(x_i)^T \phi(x_j)$ with some embedding space $\phi(\cdot)$. Then the output of LLM, $P(y | [S_n, x_{test}])$, converges polynomially to the following Kernel Regression model with probability $1 - \delta$.
    \begin{equation}
    \hat{y}_i = (\sum\nolimits_{j} y_i K(x_i, x_j)) / (\sum\nolimits_{j}K(x_i, x_j)), 
    \label{eq:before}
    \end{equation}
\end{theorem}
Theorem~\ref{theorem:icl}~\citep{han2023context} suggests that, when LLM is prompted with a set of demos and a new test query ($x^{\text{test}}$), demos that are semantically closer to the test example in embedding space contribute more to its prediction.
The same phenomenon has been observed by several empirical studies~\cite{rubin2021learning, han2023context, liu2021makes}.
This behavior is also intuitive for ICL: the whole purpose of providing demos is for LLMs to leverage and apply their patterns to the test input.

\paragraph{Clustering demos to each expert based on their semantic similarity.}
The above analysis motivates a natural way of assigning demos to different experts: by clustering them based on semantic similarity.
Starting from the kernel model in Theorem~\ref{theorem:icl}, our goal is to divide the demos into $C$ groups $\{\mathcal{V}_1, \dots, \mathcal{V}_C\}$ such that each group (expert) only uses its own demo. In this case, the same sample $x_i$'s prediction, assuming its in group $c$, will become
\begin{equation}
\bar{y}_i = (\sum\nolimits_{j\in \mathcal{V}_c} y_i K(x_i, x_j)) / (\sum\nolimits_{j\in \mathcal{V}_c}K(x_i, x_j)), 
\label{eq:after}
\end{equation}
and the error $|\bar{y}_i - \hat{y}_i|$ is related to the sum of the kernel entries outside the cluster $\sum_{j\notin \mathcal{V}_c} K(x_i, x_j)$. Therefore, a good demo assignment algorithm will minimize the sum of between-cluster kernel values while keeping the clusters balanced, leading to the following clustering objective: 
\begin{equation}
\min_{\{\mathcal{V}_1, \dots, \mathcal{V}_C\}} \sum_{c=1}^C 
\frac{\sum_{i\in \mathcal{V}_c}\sum_{j \notin \mathcal{V}_c} K(x_i, x_j)}{|\mathcal{V}_c|}.
\label{eq:minimize}
\end{equation}
Based on the derivation in \Cref{appendix:kmeans}, this is equivalent to the following clustering objective:
\begin{equation}
\min_{\{\mathcal{V}_1, \dots, \mathcal{V}_C\}} \sum_{c=1}^C \sum_{i\in \mathcal{V}_c} \|\phi(x_i) - m_c\|^2,  \ \ \ m_c = \frac{1}{|\mathcal{V}_c|}\sum_{j\in \mathcal{V}_c}\phi(x_j), 
\label{eq:K-means}
\end{equation}
which is exactly the objective function of K-means clustering in the embedding space $\phi(\cdot)$. In practice, we assume $\phi(\cdot):=\mathcal{E}_{\theta}(\cdot)$ is a mapping formed by a neural network encoder, and conduct K-means in such embedding space to cluster demos. Note that the choice of embedding space does not have to be the same as the API model; as long as the embedding space reflects the high-level semantic similarity between different demos, it can be used to effectively partition the problem space.
We also compare other options in the ablation study.

\paragraph{On the choice of clustering algorithm.}
In principle, our demo assignment method allows any clustering algorithm to be applied.
In practice, we resort to the widely adopted K-means family for their simplicity, and made the following changes to better suit our application:
1) We select the K-means-auto variant as it can infer the optimal number of experts from the data.
2) To avoid biasing towards a larger number of clusters, we employ scaled inertia (\Cref{eq:scaled_inertia}) as the criterion when identifying the optimal number of experts.
\begin{sizeddisplay}{\small}
\begin{align}
   &C^* = \argmin_{C = C_{\text{min}}, \ldots, C_{\text{max}}} \left(\min_{\{\mathcal{V}_1, \dots, \mathcal{V}_C\}} \sum_{c=1}^C \sum_{i\in \mathcal{V}_c} \|\phi(x_i) - m_c\|^2 + \alpha C\right)
\label{eq:scaled_inertia}
\end{align}
\end{sizeddisplay}

\paragraph{Routing function.}
During inference time, each new query $x$ will be routed to its closest expert in the embedding space.
\begin{equation}
    c(x) = \argmin_{c = 1, \ldots, C^*} K\left({\phi_{\theta}(x), \bm{\mu}_{c} }\right)
\label{eq:routing}
\end{equation}
Here, $\bm{\mu}_{c}$ is the clustering centroids for each expert.
The assigned expert $c(x)$ will then use its prompt (instruction + demos) to make the final prediction.


\subsection{Instruction Assignment}
\label{sec:method-inst_assign}

Given the set of demos assigned to each expert, the final step is to identify the best instruction for each cluster, so that the collective performance of the mixture is maximized.
We introduce a \textbf{Region-Based Joint Search (RBJS)} algorithm for solving this objective.
RBJS consists of two parts: generating candidate instructions and identifying the best one for each expert.

\paragraph{Generating candidate instructions to complement the demos.}
As discussed in Section~\ref{sec:method-overview}, each expert acquires a different specialty from the local information stored in their assigned demos.
Because of this, they also process distinct blind spots in terms of their general task-solving ability.
Therefore, they might require different instructions to compensate for their special needs.
Inspired by this analysis, for each expert, we propose to generate candidate instructions utilizing the demos assigned to other experts.
This way, the instruction can potentially capture the missed information.

This choice is also supported by the empirical risk minimization framework, as outlined in \Cref{sec:prelim}.
For prompt optimization, 'demons' — analogous to training data in supervised learning — incorporate all task-relevant external information accessible to the model.
Therefore, utilizing each expert’s local demonstrations for instruction generation merely duplicates the existing information.
This contradicts our goal of creating instructions that compensate for the unique specialties of each expert.
 
Any existing instruction generation algorithm can be used to propose candidate instructions given a set of demos.
In this work, we choose APE for its simplicity.

\paragraph{Identifying the best candidate for each expert.}
To select the best instruction from a set of proposals, existing prompt optimization algorithms commonly rank their performance on a held-out validation set, sampled from the same distribution as the training demos.
Using the entire validation set measures how well an expert (instruction, demos) performs on the full data distribution.
However, this might not serve our purpose: During inference, each expert is only responsible for predicting the data within their region.
Our empirical results also support this analysis; we find that the performance of an expert between the full and local data distribution is not necessarily aligned (\Cref{fig:rbjs}).

To alleviate the issue in an exhaustive search, we first route each input in the validation set to its experts, then perform a joint search on the optimal (instruction, demos) pair.

Algorithm~\ref{algo:mop} in the appendix summarizes the entire search process of the Region-Based Joint Search algorithm.

%% file: figures_tex/mop.tex
\begin{figure*}[t!]
\centering
\includegraphics[width=0.98\linewidth]{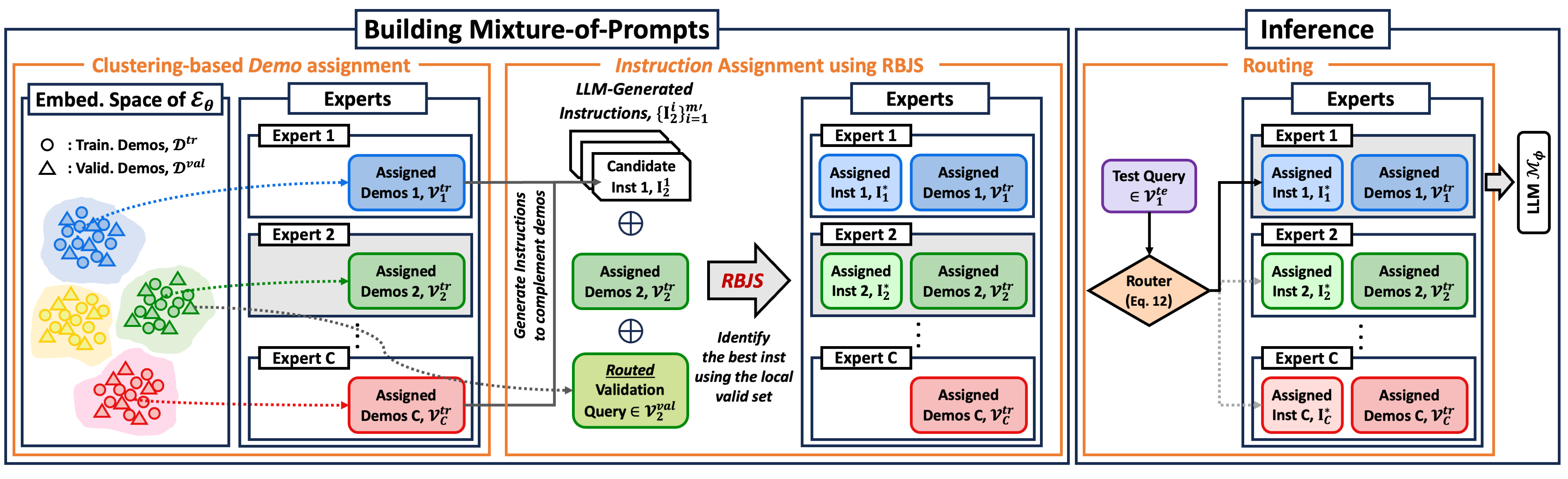}
\caption{\textbf{Illustration of MoP.} We adopt the MoE paradigm and divide the problem space into a set of sub-regions. Each sub-region is governed by a specialized expert, equipped with both an instruction and a set of demos. A two-phase process is developed to construct the specialized expert for each region: (1) \textit{demo assignment}: Inspired by the theoretical connection between ICL and kernel regression, we group demos into experts based on their semantic similarity; (2) \textit{instruction assignment}: A region-based joint search (RBJS) of an instruction per expert complements the demos assigned to it, yielding a synergistic effect. During inference, each new query is routed to its closest expert in the embedding space and the assigned expert then utilizes its prompt (instruction + demos) to make the final prediction.}
\label{fig:mop}
\end{figure*}

%% file: 5_experiment.tex
\section{Experiments}
\label{sec:exp}
 
In this section, we experimentally validate MoP, which jointly searches for the optimal 
 (instruction, demos) pair to partition the problem space into homogeneous regions. 



\input{figures_tex/analysis}
\subsection{Experimental Setup}
\label{sec:exp:setting}

\paragraph{Settings.}
We follow the settings in the original APE paper~\citep{zhou2022large} with the following exceptions.
(1) Our evaluation is conducted on OpenAI's latest GPT-3.5-Turbo-Instruct model \footnote{https://platform.openai.com/docs/models/gpt-3-5}, a cost-efficient (100$\times$ cheaper) replacement for the text-davinci model used in APE. We reran APE on this model.
(2) For all our methods, we report the mean and standard deviation across 3 runs to account for the randomness in the search phase.

\paragraph{Tasks and Evaluation Metrics.}
We empirically validate the strength of MoP across three major prompt optimization benchmarks: \textbf{Instruction Induction}~\citep{ii}, \textbf{Super Natural Instructions}~\citep{superni}, \textbf{BIG-Bench-Hard}~\citep{bbh}. These benchmarks cover a wide range of possible tasks, including coding, math, common-sense reasoning, knowledge retrieval, etc. We provide detailed descriptions of each task in the \Cref{appendix:tasks}. For these benchmarks, we measure task performance for each task using predefined score functions (details in \Cref{appendix:score_functions}). 

\paragraph{Baselines.}
We compare our method against following baselines: \textbf{APE}~\citep{zhou2022large} searches for a single instruction among a pool of instruction candidates proposed by a LLM, \textbf{APE+Demos} combines randomly selected demos with an APE-found instruction, \textbf{APE+K-centroids} combines demos corresponding to the centroids of K-means with an APE-found instruction, \textbf{InstructZero; IZ}~\citep{chen2023instructzero} finds a single instruction for a black-box LLM by optimizing the soft prompt of an open-source LLM using a Bayesian Optimization approach, and \textbf{IZ+Demos} and \textbf{IZ+K-centroids} are the same as the previous APE variants. For more details, refer to \Cref{app:baselines}. We defer more baselines that partially use our method to the ablation study in~\Cref{sec:ablate}.

\paragraph{Implementation Details.}
Following the previous automatic prompting works~\citep{zhou2022large}, we set the temperature to 0.9 when generating instructions using LLMs to encourage diversity and to 0.0 when evaluating with LLMs. 
Furthermore, for a fair comparison, \textbf{we set the same budget for all methods}. Regarding demo assignments in MoP, we use the default hyperparameter consistently across all experiments.
More details can be found in \Cref{app:imple_details}.

\subsection{Analysis}
\label{sec:exp:analysis}
Before delving into the main experiments, we conduct an empirical analysis that motivates the development of our MoP framework. The results presented here are obtained for the Auto categorization task, with the maximum number of experts set to four.

\paragraph{Visualization of demo clusters.}
Building upon the theoretical connection between ICL and Kernel Regression, we begin by clustering a given set of demos into regions based on their semantic similarity. To achieve this, we first map the given demo sets into the embedding space using text-embedding-ada-002 model \footnote{https://openai.com/blog/new-and-improved-embedding-model} as a text encoder ($\mathcal{E}_{\theta}$), and then apply the clustering algorithm described in \Cref{sec:method-demo_assign}. \Cref{fig:demo_clusters} visualizes clustering in the embedding space with t-SNE projection. The illustration indicates that \textbf{there exist underlying patterns in the data distribution, and demos with semantically similar meanings are grouped closely}. For example, for the 'Auto categorization' task shown in \Cref{fig:demo_clusters}, demos relevant to country, computer science, extinct languages, and apparel are each clustered together. By clustering demos in the embedding space, we can effectively find semantically similar clusters that help allocate test queries (marked with stars \Cref{fig:demo_clusters}) accurately to the corresponding region and the optimal expert.

\paragraph{Experts process different specialties.}
We then verify the impact of demo clusters on performance for each test query. In order to eliminate the impact of instructions on performance, all experts utilize only clustered demos as prompts, thereby restricting the output space to demos only. Subsequently, we calculate the Hit Ratio by counting the number of correctly answered experts out of the total number of experts ($C$) for each test input. If test inputs yield Hit Ratios within the range other than 0/$C$ and $C$/$C$, it indicates their sensitivity to the assigned expert.
As depicted in~\Cref{fig:diff_st}, we measure the Hit Ratios and observe that 83\% of test inputs have Hit Ratio values that are neither 0 nor 1. This implies that most test inputs are influenced by the type of clustered demos they are assigned; \textbf{Each expert develops distinct specialties based on the local information within their assigned demos, resulting in distinct blind spots in terms of task-solving ability for each test input.}

\paragraph{Necessity of RBJS.}
We also examine the necessity of jointly searching for the optimal (instructions, demos) pair. To investigate this, we initially assign training demos to 4 different experts and consider 8 candidate instructions generated by a LLM. Then, for each expert, we vary the instructions and measure the test performance of prompts by combining these instructions with the demos specific to each expert. \Cref{fig:rank_change} visualizes the ranks of the candidate instructions for each individual expert. As depicted in \Cref{fig:rank_change}, the rankings of each candidate instruction vary significantly across different experts. 
These results indicate that \textbf{a single instruction is insufficient for all experts, highlighting the need for distinct synergistic instructions for each expert}. This emphasizes the importance of a joint optimization scheme for the (instruction, demos) pair, which can lead to improved results. 
Furthermore, for each region, we calculate the correlation between 1) the validation rankings of candidate instructions obtained from a random subset of the full validation set (Joint Search) and 2) the rankings obtained when using an equal-sized routed local validation set within the target region (RBJS).
As illustrated in \Cref{fig:rbjs}, it is evident that \textbf{there is a misalignment in the validation rankings of candidate instructions between the use of a random subset of the full dataset and the utilization of routed local data}.
We defer the evaluation of the RBJS's effectiveness to an ablation study in \Cref{sec:ablate:prompt_assign}.

\input{figures_tex/main/na_10_2}
\input{figures_tex/win_rate}

\subsection{Main Results}
\label{sec:exp:main}
In this section, we conduct a large-scale experiment to compare MoP against six previous SOTAs across three major benchmarks: Instrucion-Induction~\citep{zhou2022large}, Super Natural Instruction~\cite{superni}, and BIG-Bench-Hard~\cite{bbh}.
Tasks from these benchmarks cover a broad spectrum of language understanding scenarios, including various types of reasoning, knowledge retrieval, coding, math, etc.
Due to space limit, we only showcase 19 tasks here, and include all results in \Cref{app:main}.
As shown in \Cref{fig:main_exp}, MoP outperforms prior arts (APE + Demos and IZ + Demos) by a substantial margin.

In addition, we also compute the win rate of every pair of methods.
\textbf{As shown in \Cref{fig:win_rate}, the win rate of MoP dominates all six prior methods, with an average of 81\% across three benchmarks.
The results not only reveal the strength of our framework but also demonstrate that MoP can generalize to a wide range of tasks.}

%% file: figures_tex/analysis.tex
\begin{figure*}[t!]
    \centering

    \begin{subfigure}{0.24\linewidth}
        \centering
        \includegraphics[width=\linewidth]{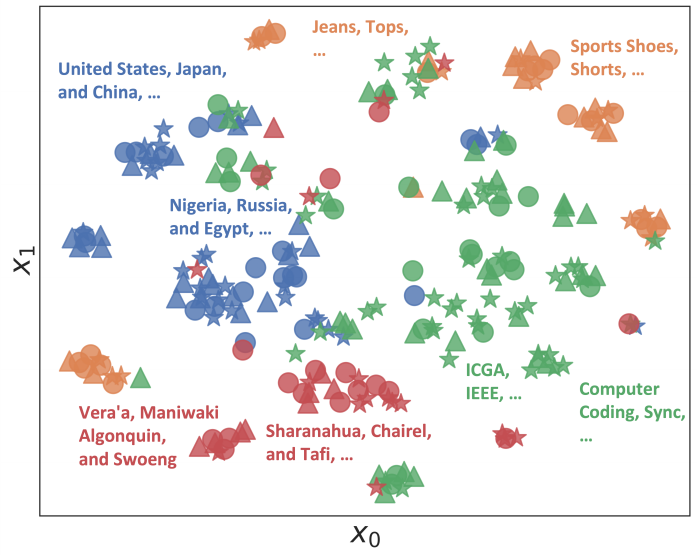}
        \caption{Visualization of demo clusters}
        \label{fig:demo_clusters}
    \end{subfigure}
    \hfill
    \begin{subfigure}{0.24\linewidth}
        \centering
        \includegraphics[width=\linewidth]{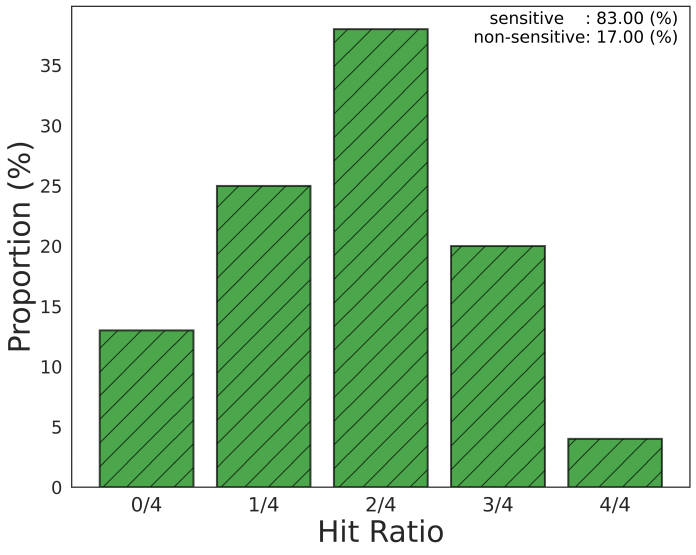}
        \caption{Experts' different strengths}
        \label{fig:diff_st}
    \end{subfigure}
    \hfill
    \begin{subfigure}{0.24\linewidth}
        \centering
        \includegraphics[width=\linewidth]{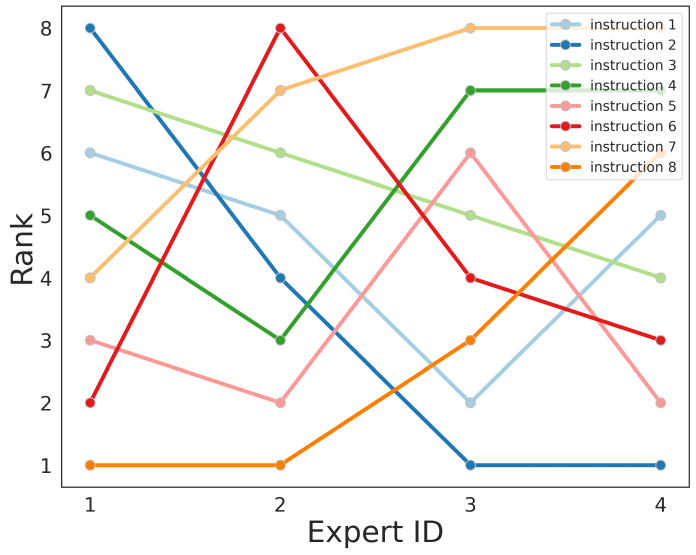}
        \caption{Necessity of Joint Search}
        \label{fig:rank_change}
    \end{subfigure}
    \hfill
    \begin{subfigure}{0.24\linewidth}
        \centering
        \includegraphics[width=\linewidth]{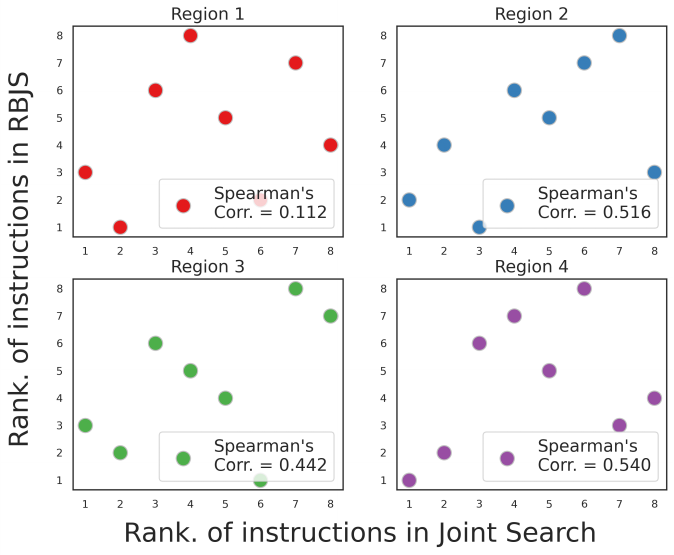}
        \caption{Necessity of RBJS}
        \label{fig:rbjs}
    \end{subfigure}
    
    \caption{\textbf{Analysis.} \textbf{(a)} There exist underlying patterns in the data distribution, and demos with semantically similar meanings are grouped closely. The circle, triangle, and star shapes represent training, routed validation, and routed test demos, respectively. \textbf{(b)} Each expert has distinct task-solving ability for each input. \textbf{(c)} A single instruction is insufficient for all experts, highlighting the need for distinct synergistic instructions for each expert. \textbf{(d)} Performance of instructions evaluated under local data distribution for each expert (i.e. subsets routed to each expert) is not aligned with the full data; This motivates performing region-based evaluation during joint search (RBJS).}
    \label{fig:analysis}
\end{figure*}

%% file: figures_tex/main/na_10_2.tex
\begin{figure*}[t!]
    \centering
    \includegraphics[width=0.835\linewidth]{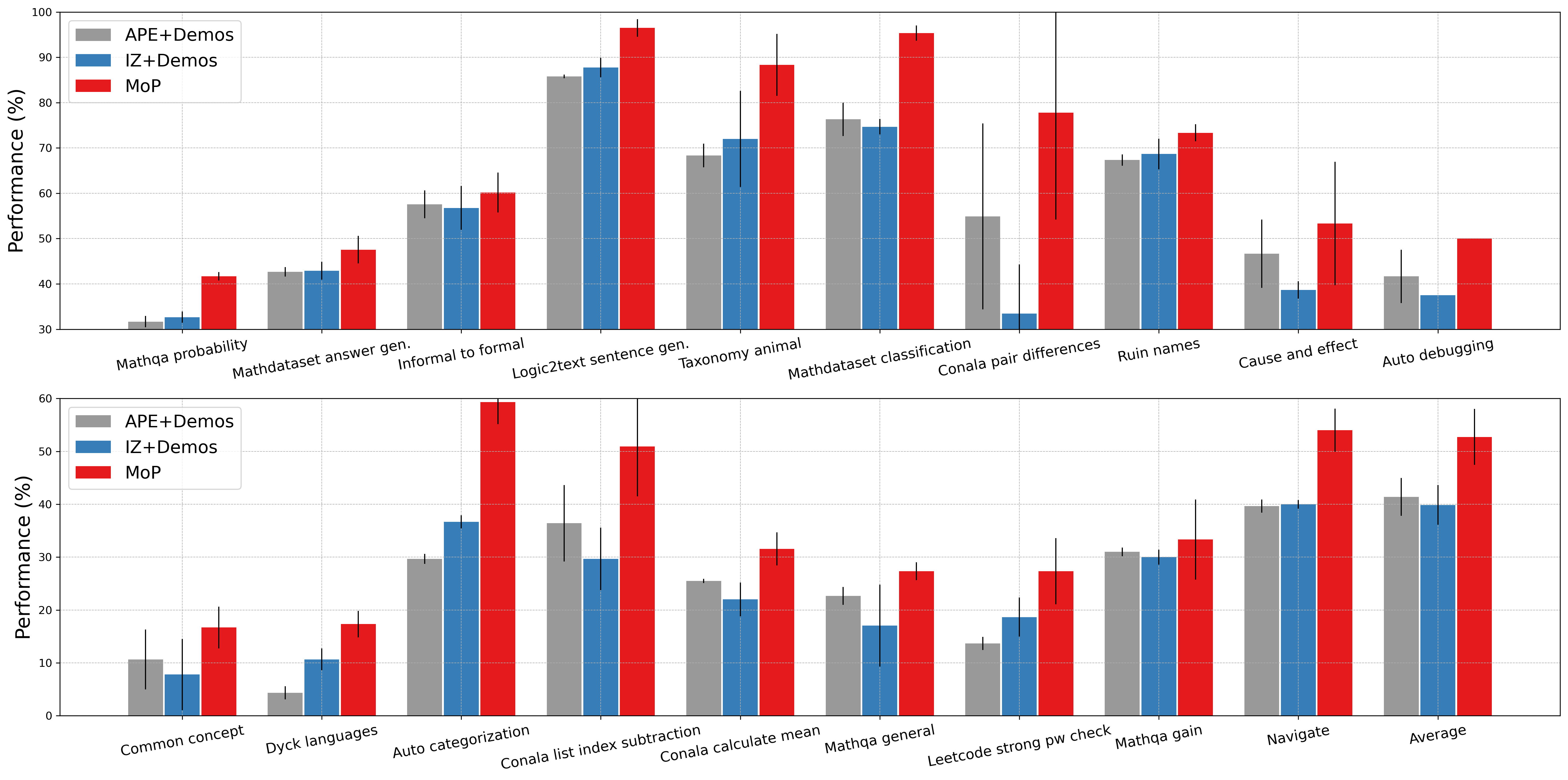}
    \caption{\textbf{Main results.}  We validate MoP across three major prompt optimization benchmarks. MoP achieves an average performance of \textbf{52.73\%} outperforming the average performance of 41.39\% / 39.87\% achieved by APE+Demos / IZ+Demos in these results.}
    \label{fig:main_exp}
\end{figure*}

%% file: figures_tex/win_rate.tex
\begin{figure*}[t!]
    \centering

    \begin{subfigure}{0.33\linewidth}
        \centering
        \includegraphics[width=\linewidth]{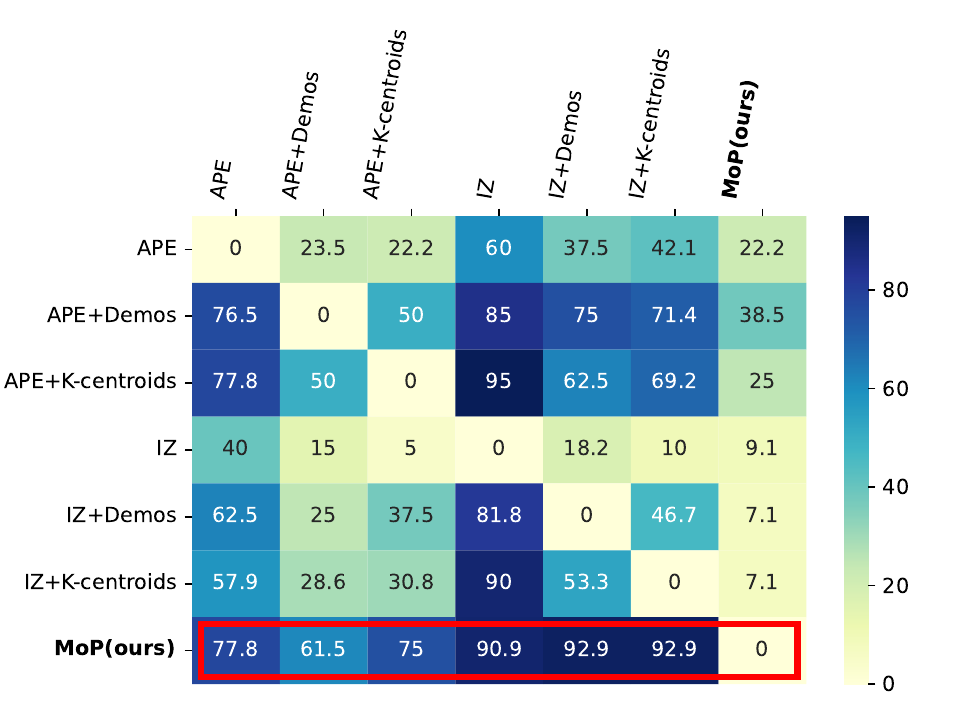}
        \caption{Instruction Induction}
        
    \end{subfigure}
    \hfill
    \begin{subfigure}{0.33\linewidth}
        \centering
        \includegraphics[width=\linewidth]{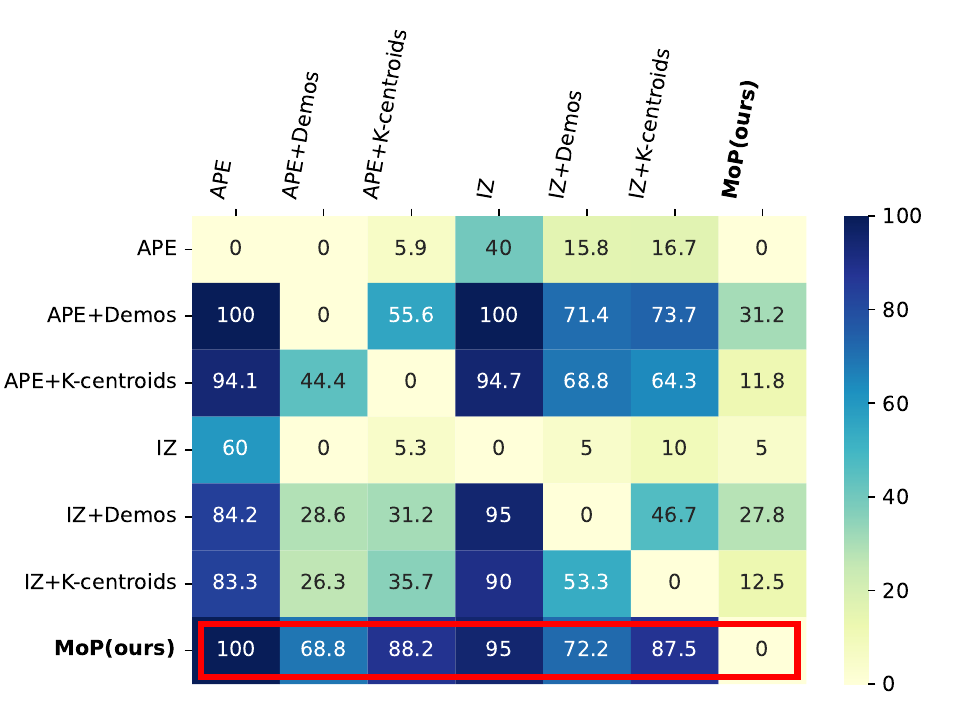}
        \caption{Super Natural Instructions}

    \end{subfigure}
    \hfill
    \begin{subfigure}{0.33\linewidth}
        \centering
        \includegraphics[width=\linewidth]{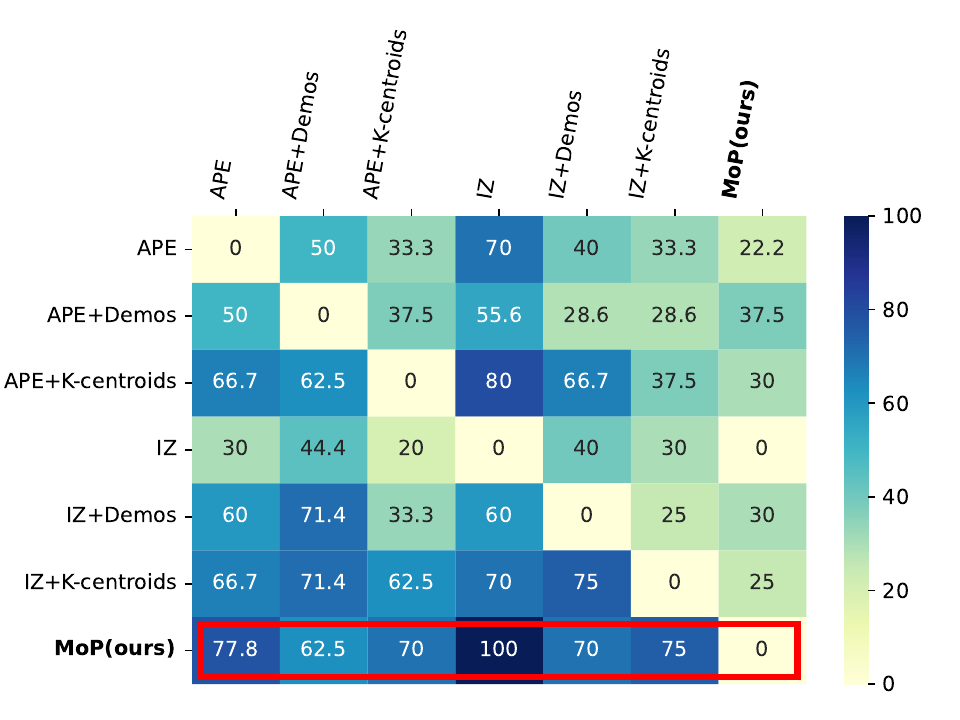}
        \caption{BIG-Bench-Hard}
        
    \end{subfigure}
    
    \caption{\textbf{Win rate matrices.} We compare the pairwise win rate of all methods on Instruction Induction (\textbf{a}), SuperNI (\textbf{b}), and BIG-Bench-Hard (\textbf{c}). Our method achieves the best win rate against all six baselines across various benchmarks. The average win rate of MoP across all benchmarks is 81\%.}
    \label{fig:win_rate}
\end{figure*}

%% file: 6_ablate.tex
\section{Ablation Study}
\label{sec:ablate}
In this section, we ablate the effect of different modules in the proposed MoP framework.
We conduct the experiments on three tasks: Auto categorization, Mathdataset classification, and Taxonomy animal the task throughout the section.
All other settings are identical to the previous section.

\subsection{Robustness of MoP on Out-of-Distribution Data}
\label{sec:ablate:ood}
\input{tables/rebuttal/ood}
To further assess the robustness of our method under Out-Of-Distribution (OOD) settings, we craft OOD datasets that can challenge MoP:
Using the same embedding model as MoP, we divided the original dataset into two clusters: one designated for training and the other for testing.
This division ensured that all test data were significantly distant from the training clusters, providing a rigorous test of MoP’s demonstration assignment and routing functions — arguably a more adversarial setup than that faced by APE.

The results in Table~\ref{sec:ablate:ood} reveal several insights. Firstly, all methods exhibited a performance drop on the OOD dataset. This aligns with the principle of empirical risk minimization, where optimization is strictly confined to the information provided by the training data.
Secondly, MoP consistently outperforms other baselines.

The resilience of MoP can be attributed to its strategic segmentation of the problem space. By dividing the space into distinct regions, each managed by an expert, MoP ensures that even an OOD query is matched to the closest region.
This reduces the "out-of-distribution" effect for the query relative to the localized data distribution, making the query effectively less alien to the selected expert region.

\subsection{Different Number of Demos}
\label{sec:ablate:num_demos}
\input{figures_tex/abl_num_demos/abl_num_demos}

Firstly, we verify the performance of MoP against baselines across different numbers of demos.
As shown in \Cref{fig:abl_ndemos}, MoP consistently outperforms the baselines across various numbers of demos, achieving a significant improvement of 16.89\%, 22.89\%, 2.78\% compared to APE+Demos and 21.67\%, 19.88\%, 3.33\% compared to IZ+Demos across all the tasks considered for each number of demos.

\subsection{Different Clustering Algorithm}
\label{sec:ablate:clustering}
We use K-means-Auto to cluster demos, which automatically decides the best number of experts.
The intuition behind it is that more experts do not necessarily produce the best result.
Here, we validate this choice by comparing the performance of K-Means-Balanced and K-means-Auto.
As shown in \Cref{tab:ablate.cluster_embed_pa}, both K-Means variants outperform random, while K-Means-Auto achieves the best results.

\subsection{Different Embedding Model}
\label{sec:ablate:embed_model}
During demo assignment, we measure the semantic similarity of demos using $l_2$ distance in the embedding space.
While our method is agnostic to the specific choice of embedding models, stronger text encoders perform better in identifying semantically similar demos.
Here we examine how the strength of the embedding model affects the performance of MoP.
We examine three commonly used text encoders: GPT2-Large, T5, and Ada-002.
The results in \Cref{tab:ablate.cluster_embed_pa}'s 1st group suggests that, while Ada achieves the best result, MoP operates reasonably well with paried with GPT2-Large.
This shows that the proposed demo assignment method does not rely on a specific embedding model.

\subsection{Different Prompt Assignment Algorithms}
\label{sec:ablate:prompt_assign}
We further ablate the key elements behind the design of our Region-Based Joint Search algorithm.
1). The optimal instruction for each expert might be distinct, therefore their generation and assignment should be conditioned on the demo clusters (Joint Search);
2). To find instructions that compensate for each expert's demos, we use demos from all other experts to generate instructions.
3). The optimal prompt for each expert is evaluated only on the text points assigned to this expert (Region-based).
We designed three prompt assignment methods to validate these hypotheses respectively: 1). \textbf{Independent Search} searches for the best prompt and assigns demos independently, i.e. the global best prompt will be assigned to all experts; 2). \textbf{RBJS Same-Cluster} uses each expert's own demos to generate prompts; 3). \textbf{Joint Search} ranks the best prompt on all validation data.
The results in the bottom group of \Cref{tab:ablate.cluster_embed_pa} confirm the claims: Region-Based Joint Search achieves the best results.

\subsection{Different Routing Algorithms}
\label{sec:ablate:routing}
The routing function is crucial to the MoE framework, as it is responsible for assigning the test input to the most suitable experts.
Following the clustering-based demo assignment, our routing function maps a test input to its closest expert in embedding space as well.
As shown in \Cref{tab:ablate.cluster_embed_pa}, our routing function significantly outperforms random assignment.

\input{tables/ablate_cluster_embed_pa}

%% file: tables/rebuttal/ood.tex
\begin{table}[h]
    \vspace{-2mm}
    \centering
    \caption{\textbf{Comparison on Out-of-Distribution data.} We compare different methods on Out-of-Distribution data, created adversarially using the embedding model (same as MoP) to group the original dataset into two clusters, one for training and one for testing. Overall, MoP achieves the best robustness under this setting.}

    \resizebox{0.95\linewidth}{!}{
        \begin{tabular}{p{3cm}ccccccc}
        
            \toprule
            \textbf{OOD Data} & \textbf{Mathdataset} & \textbf{Taxonomy animal} & \textbf{Auto cate} \\ \midrule
            \midrule
            APE                   & 9 $\pm$ 11.0 & \bf{62 $\pm$ 3.6} & 37 $\pm$ 4.1 \\
            APE + random          & 44 $\pm$ 6.6 & 61 $\pm$ 3.3 & 38 $\pm$ 0.4 \\
            APE + kcen            & 40 $\pm$ 11.8 & 60 $\pm$ 1.0 & 44 $\pm$ 2.2 \\
            IZ                    & 25 $\pm$ 26.0 & 55 $\pm$ 5.8 & 44 $\pm$ 10.1 \\
            IZ + random           & 45 $\pm$ 9.9 & 60 $\pm$ 0.6 & 43 $\pm$ 0.4 \\
            IZ + kcen             & 48 $\pm$ 11.7 & 61 $\pm$ 1.4 & 44 $\pm$ 1.2 \\
            MoP                   & \bf{68 $\pm$ 4.7} & 60 $\pm$ 1.5 & \bf{46 $\pm$ 1.8} \\
            \bottomrule
        
        \end{tabular}
    }

    \vspace{-2mm}
    \label{tab:ablate.cluster_embed_pa}
\end{table}

%% file: figures_tex/abl_num_demos/abl_num_demos.tex
\begin{figure*}[t!]
    \centering
    
    \hfill
    \begin{subfigure}{0.30\linewidth}
        \centering
        \includegraphics[width=\linewidth]{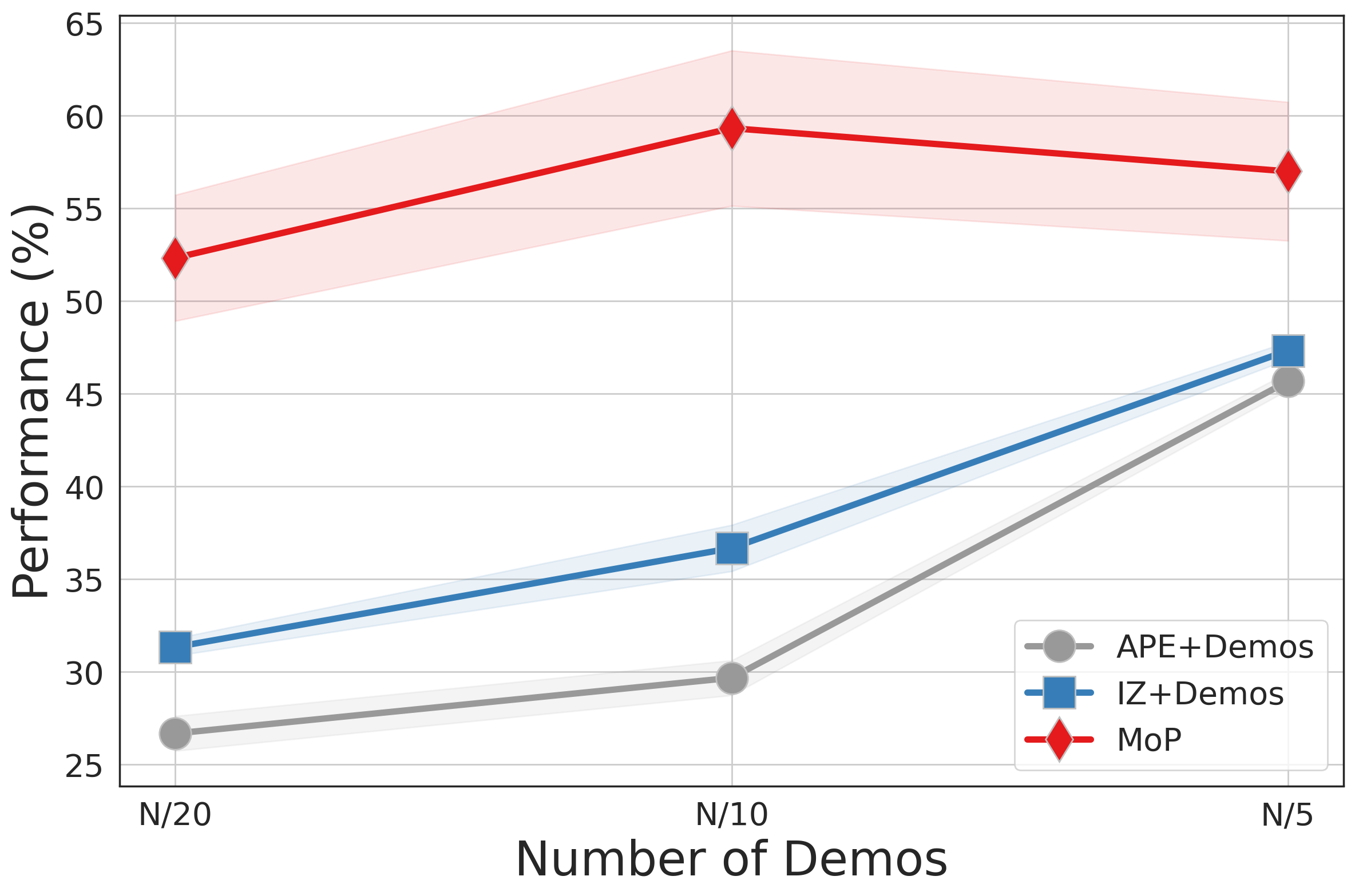}
        \caption{Auto categorization}
        
    \end{subfigure}
    \hfill
    \begin{subfigure}{0.30\linewidth}
        \centering
        \includegraphics[width=\linewidth]{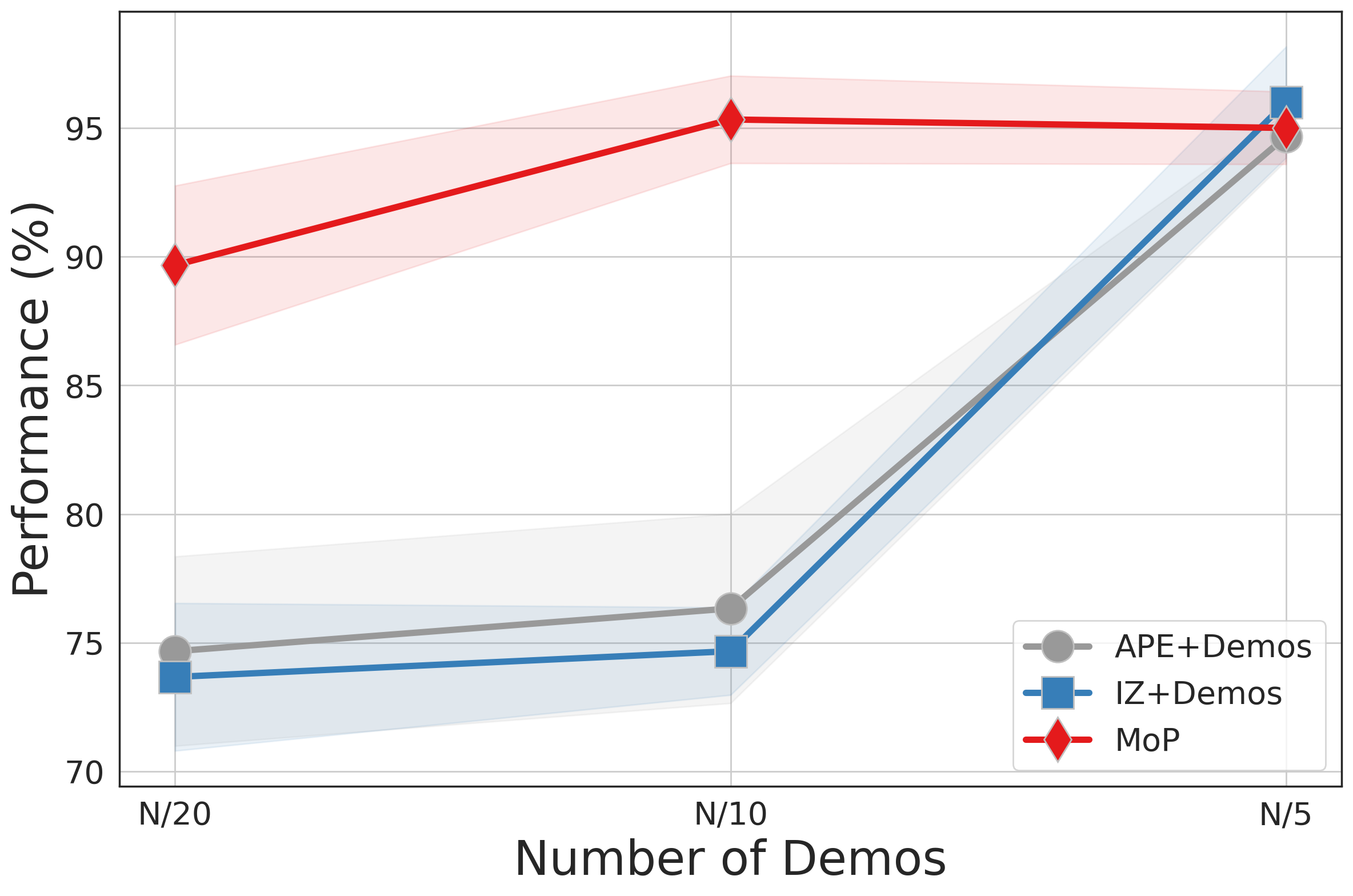}
        \caption{Mathdataset classification}
    \end{subfigure}
    \hfill
    \begin{subfigure}{0.30\linewidth}
        \centering
        \includegraphics[width=\linewidth]{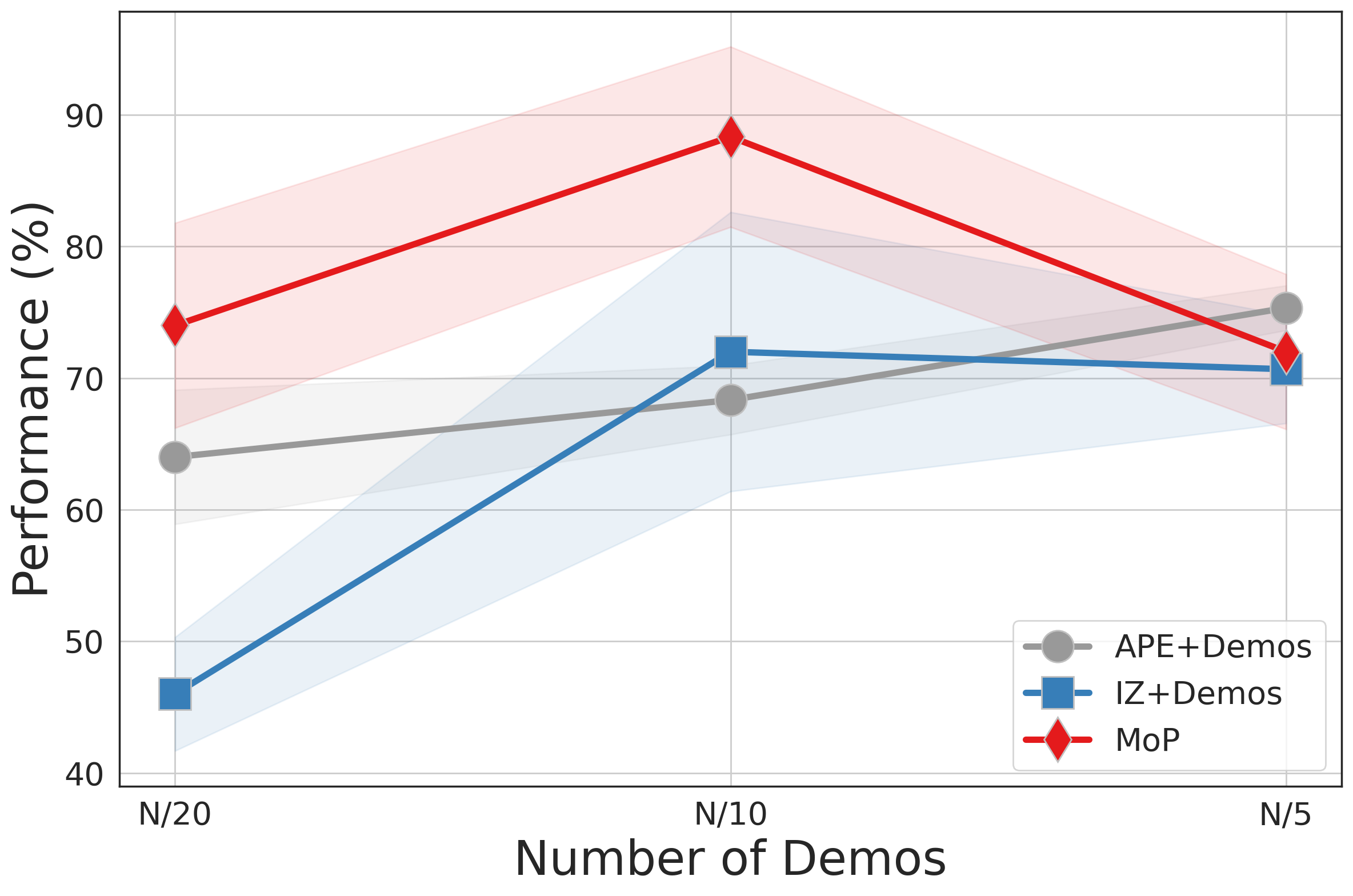}
        \caption{Taxonomy animal}
    \end{subfigure}
    \hfill

    \caption{\textbf{Ablation study on different number of demos.} We measure the task performance of each method across different numbers of demos. Here, $N$ on the x-axis represents the total number of training demos.}
    \label{fig:abl_ndemos}
\end{figure*}

%% file: tables/ablate_cluster_embed_pa.tex
\begin{table}[t]
    \vspace{-2mm}
    \centering

    \caption{\textbf{Ablation study.} The choice of models and algorithms in MoP is listed in the last row of every group. The best performance is achieved with K-means-Auto, text-embedding-ada, and RBJS.}

    \resizebox{0.95\linewidth}{!}{
        \begin{tabular}{p{3cm}ccccccc}
        
            \toprule
            \textbf{Embed Model} & \textbf{Mathdataset} & \textbf{Taxonomy animal} & \textbf{Auto cate} \\ \midrule \midrule
            GPT2-Large  & \bf{97 $\pm$ 1.3} & \bf{89 $\pm$ 0.9} & 53 $\pm$ 0.9 \\ 
            Sentence-T5 & 92 $\pm$ 2.3 & 76 $\pm$ 3.9 & 53 $\pm$ 5.3 \\
            Ada         & 95 $\pm$ 3.3 & \bf{88 $\pm$ 6.9} & \bf{59 $\pm$ 0.4} \\ 
            \bottomrule
        
        \end{tabular}

    }

    \resizebox{0.95\linewidth}{!}{
        \begin{tabular}{p{3cm}ccccccc}
        
            \toprule
            \textbf{Clustering} & \textbf{Mathdataset} & \textbf{Taxonomy animal} & \textbf{Auto cate} \\ \midrule \midrule
            Random              & 89 $\pm$ 1.3 & 74 $\pm$ 5.7 & 36 $\pm$ 0.8 \\ 
            K-means-Balanced    & \bf{96 $\pm$ 0.8} & 82 $\pm$ 2.6 & 52 $\pm$ 2.6 \\ 
            K-means-Auto        & \bf{95 $\pm$ 1.7} & \bf{88 $\pm$ 6.9} & \bf{59 $\pm$ 4.2} \\
            \bottomrule

        \end{tabular}
    }

    \resizebox{0.95\linewidth}{!}{
        \begin{tabular}{p{3cm}ccccccc}
        
            \toprule
            \textbf{Prompt Assignment} & \textbf{Mathdataset} & \textbf{Taxonomy animal} & \textbf{Auto cate} \\ \midrule \midrule
            Independent Search  & 94 $\pm$ 1.5 & 71 $\pm$ 6.3 & 52 $\pm$ 3.4 \\
            Joint Search (JS)   & 93 $\pm$ 1.7 & 82 $\pm$ 9.0 & 56 $\pm$ 2.9 \\
            RBJS Same-Cluster   & 91 $\pm$ 1.4 & 72 $\pm$ 1.0 & \bf{60 $\pm$ 4.9} \\
            RBJS (Ours)         & \bf{95 $\pm$ 1.7} & \bf{88 $\pm$ 6.9} & \bf{59 $\pm$ 4.2} \\
            \bottomrule
            
        \end{tabular}
    }

    \resizebox{0.95\linewidth}{!}{
        \begin{tabular}{p{3cm}ccccccc}
        
            \toprule
            \textbf{Routing Function} & \textbf{Mathdataset} & \textbf{Taxonomy animal} & \textbf{Auto cate} \\ \midrule \midrule
            Random   & 77 $\pm$ 2.2 & 87 $\pm$ 3.3 & 32 $\pm$ 2.9 \\
            MoP      & \bf{95 $\pm$ 1.7} & \bf{88 $\pm$ 6.9} & \bf{59 $\pm$ 4.2} \\
            \bottomrule
        
        \end{tabular}
    }

    \label{tab:ablate.cluster_embed_pa}
    \vspace{-3mm}
\end{table}

%% file: 7_conclusion.tex
\section{Conclusion}
\label{sec:conclusion}


This work introduces the Mixture-of-Prompts (MoP) approach to enhance the performance of prompt optimization for Large Language Models.
While existing methods search for a single instruction to prompt language models, MoP optimizes for a set of experts (prompts), each governing a specialized region of the problem space.
This divide-and-conquer approach reduces the complexity associated with the task assigned to a single prompt, thereby substantially enlarging the problem space coverage.
Within MoP framework, we further investigate various demo and instruction assignment methods for constructing the expert committee.
Equipped with the proposed similarity-based demo assignment and region-based demo-instruction joint search, MoP substantially improves the performance over comparable methods over a diverse set of NLP tasks.
We hope the proposed method and associated findings could open up new possibilities for prompt optimization research.

\paragraph{Limitations}
We include a discussion on the limitations of our method in \Cref{app:limitation}.

\section*{Acknowledgements}
The work is partially supported by NSF 2048280, 2331966, 2325121, 2244760, ONR N00014-23-1-2300, Institute for Information \& communications Technology Promotion(IITP) grant funded by the Korea government(MSIT) (No.RS-2019-II190075 Artificial Intelligence Graduate School Program(KAIST), and the National Research Foundation of Korea(NRF) grant funded by the Korea government(MSIT) (No. RS-2023-00256259).

\section*{Impact Statement}
In this work, we validate that our framework enhances the performance of prompt optimization for Large Language Models, by introducing the Mixture-of-Prompts, as shown in \Cref{sec:exp}. Our approach potentially accelerates the effective utilization of Large Language Models, which have consistently showcased remarkable generalization capabilities across diverse tasks, contributing to promoting more equitable access to technological resources. Nevertheless, it is essential to acknowledge the possibility of our framework being misused for purposes such as generating misleading information or promoting unethical practices. We hope that our research will be applied responsibly and ethically.

%% file: appendix.tex
\newpage
\appendix
\onecolumn
\begin{center}{\bf {\LARGE Appendix}}\end{center}
\paragraph{Organization} The appendix file is organized as follows:
\begin{itemize}
    \item \textbf{\Cref{appendix:baselines}} - We provide comparison with additional baselines: APE-Nearest Neighbor and OPRO.
    \item \textbf{\Cref{appendix:algo_mop}} - We provide the algorithm of the proposed approach, MoP.
    \item \textbf{\Cref{appendix:moe_theory}} - We provide a theoretical connection between MoP and MoE.
    \item \textbf{\Cref{appendix:search_cost}} - We provide a comparison of the runtime associated with different methods.
    \item \textbf{\Cref{appendix:example}} - We provide representative examples for both success and failure cases.
    \item \textbf{\Cref{appendix:kmeans}} - We provide the derivation of clustering objective.
    \item \textbf{\Cref{appendix:template}} - We provide templates used in each scenario in our experiments.
    \item \textbf{\Cref{appendix:tasks}} - We provide detailed descriptions for each task.
    \item \textbf{\Cref{appendix:score_functions}} - We provide an explanation of the metrics used for evaluating prompts.
    \item \textbf{\Cref{app:main}} - We provide the entire results for the main experiment alongside \Cref{sec:exp:main}.
    \item \textbf{\Cref{app:baselines}} - We provide further descriptions of the baselines.
    \item \textbf{\Cref{app:imple_details}} - We provide additional descriptions of the implementation details for the experimental settings.
    \item \textbf{\Cref{app:limitation}} - We conclude by outlining the limitations of our work.    
\end{itemize}

\section{Comparison with additional baselines}
\label{appendix:baselines}
\input{tables/rebuttal/more_baseline_bbh}
We conducted further experiments to compare our method with two additional baselines:
\textbf{(1) OPRO}~\citep{opro} - a recently proposed genetic algorithm-based prompt optimization method.
\textbf{(2) APE + Nearest Neighbor}: At test time, we select demonstrations based on their proximity to the test query.
We focus on the Big-Bench-Hard for those experiments, as it contains some of the hardest tasks that can stress test how each algorithm handles complex problem spaces.
For these experiments, we focus on the BBH benchmark, known for its challenging tasks, to assess how each algorithm performs in complex problem-solving scenarios.

\paragraph{OPRO}
Since OPRO only provides implementation for BBH and its best results are obtained using powerful proprietary LLMs, we rerun OPRO with GPT-3.5-Turbo for fair comparison.
As summarized in ~\Cref{tbl:more_baseline_bbh}, our method (MoP) achieves an 80\% win rate against this newer prompt optimization approach.

\paragraph{APE + Nearest Neighbor}
We employ the same embedding model, \texttt{text-embedding-ada-002}, as used in MoP for the distance function in the nearest neighbor search.
The results, presented in ~\Cref{tbl:more_baseline_bbh}, show that MoP achieves a win rate of 70\% over APE + Nearest Neighbor.
This result provides further evidence of the necessity of jointly optimizing instructions and demonstrations:
Since the Nearest Neighbor demo set can only be determined at inference time, it relies on an independently searched demo-free instruction (i.e., APE/IZ), which results in a suboptimal combination.
In contrast, for MoP, prompts and demonstrations are jointly optimized for each expert, creating a coherent skill set. Each expert is initially assigned a fixed cluster of demonstrations; subsequently, the prompt assignment algorithm selects the best instruction for each expert separately, to enhance the utility of their specific demonstrations.

\section{Algorithms for MoP}
\label{appendix:algo_mop}
\input{algos/mop}
\section{Theoretical connection between MoP and MoE}
\label{appendix:moe_theory}
Our work adapts the MoE framework, traditionally involving distinct models as experts, by defining experts as diverse prompts (instructions + demonstrations).
We offer the following insights to highlight the duality between this application and traditional MoE.
\begin{enumerate}
    \item \textbf{Prompt Optimization can be viewed as model selection:} An LLM pre-trained on the next-token prediction task can be seen as a collection of conditional probabilistic models, each defined by a specific prompt. By crafting various prompts, LLM users are essentially picking different submodels to perform various tasks. Thus, designing varied prompts is equivalent to selecting different sub-models from this collection, making the process of automatically optimizing the prompt for each expert parallel to constructing a suitable model for each expert in traditional MoE.
    \item \textbf{Theoretical properties of MoE apply to MoP:} This aforementioned conceptual framework allows us to directly apply the theoretical properties of MoE to the task of prompt optimization for LLMs, as optimizing for a mixture of expert prompts is identical to optimizing a mixture of expert models.
\end{enumerate}

We view the main contribution of our work as the first to adapt the MoE framework to prompt optimization tasks and bring the community's attention to its strong and consistent potential despite the choices of simple algorithms for each component (demo assignment, prompt assignment, and routing function). We hope this explanation better highlights the theoretical duality between MoP and MoE, and further motivates our work.

\section{Search cost comparison}
\label{appendix:search_cost}
\input{tables/rebuttal/search_cost}
We benchmark the runtime of the search and inference phase of MoP with different baselines.
There exist two computational components in our method: between two types of computations: 1). query LLM 2). running the embedding model.
Since the first former dominates latter in practice, we will focus on analyzing the complexity w.r.t. LLM queries:
\begin{itemize}
  \item \textbf{Inference:} MoP, operating under the Mixture of Experts (MoE) paradigm, deploys a single prompt akin to deploying a single instruction. This constitutes one key benefit of MoP (MoE) paradigm over prompt ensemble methods, or simply using longer prompts.
  \item \textbf{Search:} The search in MoP involves negligible costs for the demo assignment phase as they do not require LLM querying. For the prompt assignment phase, the complexity is linear w.r.t. the number of experts but is fully parallelizable.
\end{itemize}
Table~\ref{tab:search_cost} reports the wallclock times comparing MoP with APE and IZ.
Our findings include: (1). gs include: (1) All methods exhibit similar inference times, which aligns with our previous analysis. (2). Both APE and MoP have substantially lower search costs compared to IZ, which incurs additional costs due to the local operation of an open-sourced LLM. (3). While MoP's search cost (with 10 experts) approximately triples that of APE, this can be significantly reduced through parallelization.

\section{Qualitative analysis of the discovered experts}
\label{appendix:example}
We provide an example analysis of the discovered experts, focusing on why MoP are more (less) effective for certain tasks.

\paragraph{Example success case:}
An example where MoP significantly outperforms random demos is in the task of Auto-categorization.
This task uses training datasets with various categorization questions belonging to different genres, such as Country (e.g., countries with large populations, countries in the UN), Celebrity, Language, and Companies.
We found that each identified expert specializes in one or two categories.
For instance, one expert handles only celebrity-related queries, enhancing their ability to provide accurate answers.
Another case where MoP excels is the Movie recommendation task from BBH.
Here, each expert identified by the MoP algorithm focuses on a distinct set of movies.
For example, expert 1 focuses on classic adventures in fantasy and whimsical settings, like 'The Wizard of Oz' and 'Raiders of the Lost Ark'; while expert 5 handles movies that involve deep themes and complex stories, such as 'The Matrix' and 'Schindler’s List'."

\paragraph{Example failure case:}
An example where MoP exhibits performance similar to random demonstrations is in the task 'Larger Animals'.
In this task, MoP performs similarly to APE-Random, indicating that using multiple experts yields no additional benefit.
Upon examining the identified experts, we find negligible differences in their specializations.
This observation is intuitive, as this task involves selecting the largest animal from a randomly sampled list of animals of varying sizes; therefore, no specialized training data is necessary to successfully accomplish the task.

\section{Derivation of clustering objective}
\label{appendix:kmeans}

From \eqref{eq:minimize} we have
\begin{align*}
&\min_{\{\mathcal{V}_1, \dots, \mathcal{V}_C\}} \sum_{c=1}^C 
\frac{\sum_{i\in \mathcal{V}_c}\sum_{j \notin \mathcal{V}_c} K(x_i, x_j)}{|\mathcal{V}_c|} \\
= &\min_{\{\mathcal{V}_1, \dots, \mathcal{V}_C\}} \sum_{c=1}^C
\sum_{i\in \mathcal{V}_c} (\frac{\sum_{j\notin \mathcal{V}_c} K(x_i, x_j)}{|\mathcal{V}_c|}) \\
= &\min_{\{\mathcal{V}_1, \dots, \mathcal{V}_C\}} \sum_{c=1}^C
\sum_{i\in \mathcal{V}_c} (\frac{ \sum_{j} K(x_i, x_j) - \sum_{j\in \mathcal{V}_c} K(x_i, x_j)}{|\mathcal{V}_c|}) \\ 
= &\min_{\{\mathcal{V}_1, \dots, \mathcal{V}_C\}} \sum_{c=1}^C
\sum_{i\in \mathcal{V}_c} (K(x_i, x_i) -  \frac{\sum_{j\in \mathcal{V}_c} K(x_i, x_j)}{|\mathcal{V}_c|}) + \text{const} \\
= &\min_{\{\mathcal{V}_1, \dots, \mathcal{V}_C\}} \sum_{c=1}^C
\sum_{i\in \mathcal{V}_c} (K(x_i, x_i) -  2\frac{\sum_{j\in \mathcal{V}_c} K(x_i, x_j)}{|\mathcal{V}_c|} + \frac{\sum_{j,k\in \mathcal{V}_c} K(x_j, x_k)}{|\mathcal{V}_c|^2} )  \\
= &\min_{\{\mathcal{V}_1, \dots, \mathcal{V}_C\}} \sum_{c=1}^C
\sum_{i\in \mathcal{V}_c} (\phi(x_i)-\frac{\sum_{j\in \mathcal{V}_c}\phi(x_j)}{|\mathcal{V}_c|})^2  \\
\end{align*}

\section{Template used in our experiments}
\label{appendix:template}
Referring to \citet{zhou2022large}, we provide templates used in each scenario in our experiments. \textit{Generating instructions} refers to generating instructions, while \textit{Evaluation} denotes the inference time (validation or test phase). For the case of \textit{Listing Demos}, it refers to the template used when listing multiple demo samples. When a prompt is injected into the model, the $<$COMPLETE$>$ part is removed, and the model generates an output. For a fair comparison, the same template was applied to all methods.
\input{figures_tex/template}

\section{Tasks}
\label{appendix:tasks}
In this section, we provide detailed descriptions for each task across three benchmarks, encompassing a wide range of possible tasks, including coding, mathematics, common-sense reasoning, and knowledge retrieval: \textbf{Instruction Induction} (\Cref{tab:ape_benchmark_description}), \textbf{Super Natural Instructions for coding and mathematics} (\Cref{tab:superni_task_desc_code} and \Cref{tab:superni_task_desc_math}), and \textbf{BIG-Bench-Hard} (\Cref{tab:bbh_task_desc}).

\input{tables/rebuttal/ape_benchmark_description}

\input{tables/superni/task_desc_code}
\input{tables/superni/task_desc_math}

\input{tables/bbh/task_desc}

\newpage
\section{Score Functions}
\label{appendix:score_functions}
For the Instruction Induction benchmark tasks, we evaluate the quality of prompts using a metric called execution accuracy proposed by \citet{ii}. The metric is defined as follows: For each (input, output) pair, if the model's prediction matches the output exactly, it equals 1. If there is no perfect match, it equals 0.
In certain tasks, a modified version of this metric is employed. For instance, it measures the proportion of correct answers within the total answer set. Please refer to Section 4.2 of \citet{ii} for further details.

For the tasks in the Super Natural Instructions benchmark, we employ ROUGE-L scores as the evaluation metric, as outlined in \citet{superni}. 

For the BIG-Bench-Hard benchmark task, we utilize execution accuracy as the evaluation metric, following \citet{bbh}.





\section{Main Results}
\label{app:main}
\subsection{Results on Instruction Induction}
\label{app:main_ii}
\input{tables/ii/main_ii}
We show the execution accuracy results for each method in the entire Instruction Induction benchmark~\citep{ii} tasks in \Cref{tbl:main_ii}, excluding the two tasks for which the dataset has not been made publicly available: "Ascii" and "Cs algorithms". 

\subsection{Results on Super Natural Instructions}
\label{app:main_superni}
\input{tables/superni/task_sel_criteria}
\input{tables/superni/main_superni}
To further enhance the practical applicability of our approach, we conducted experiments on the Super Natural Instructions benchmark~\citep{wang2022super}. This benchmark encompasses a variety of tasks, including those related to commonsense classification and information extraction. Although it covers a wide range of tasks, our validation specifically focused on tasks related to code and mathematics.

To accomplish this, we began by evaluating the performance of APE on tasks related to code and mathematics. Subsequently, we conducted experiments on 20 tasks where APE encountered challenges, i.e., tasks for which APE's ROUGE-L score was below 50\% (please refer to \Cref{tbl:superni_task_sel_criteria}).

\subsection{Results on BIG-Bench-Hard benchmark}
\label{appendix:main_bbh}
\input{tables/bbh/main_bbh}
We conduct experiments on the tasks included in the BIG-Bench-Hard benchmark, which focuses on tasks believed to be challenging, among the BIG-Bench Instruction Induction tasks proposed in \citet{zhou2022large}.


\section{Baselines}
\label{app:baselines}

\subsection{APE}
\label{appendix:sub:backgrouds_ape}
In this section, to facilitate readers' understanding, we provide a detailed explanation of APE (Automatic Prompt Engineering~\citep{zhou2022large}), which is closely relevant to our work. 

\subsubsection{The background behind Automatic prompt optimization}
To begin with, we aim to explain the background behind auto-prompting methods, including APE~\citep{zhou2022large}. While recent LLMs have demonstrated their remarkable ability to solve tasks described by user instructions~\citep{chatgpt, gpt4, llama, peters2018dissecting, devlin2018bert, brown2020language, wei2022chain}, carefully crafted prompts are crucial for maximizing LLMs' problem-solving ability. However, this often involves laborious trial and error. Recent attempts automate this by using LLMs to design prompts with their language generation ability, addressing tasks given demo datasets. APE is one of these automatic prompt optimization methods, which has empirically demonstrated that LLM-generated prompts are more effective than human-crated prompts in solving target tasks.

\subsubsection{Detailed Explanation of the APE algorithm}
\label{appendix:sub:sub:algo_ape}
\input{algos/ape}

We provide a more detailed explanation of the APE method. In APE~\citep{zhou2022large}, firstly, it leverages a pre-trained black-box LLM to propose a set of candidate instructions. Specifically, APE initially selects random demos utilized for proposing instructions and adopts the templates corresponding to 'Generating Instructions' from~\Cref{fig:template}, along with the sampled demos, into [FULL\_DEMOS]. It then feeds this prompt into LLM to generate a set of candidate instructions.
After generating a set of candidate instructions in this manner, APE evaluates these generated candidate instructions using the subset of validation set ($\Tilde{\mathcal{D}}^{\text{valid}}$). Subsequently, it utilizes the best instruction with the highest validation score, which is a single demo-free instruction, during the test phase.
For a fair comparison, all methods, including our MoP method, use the same training, validation, and test datasets.

\subsection{InstructZero}
InstructZero~\citep{chen2023instructzero} finds a single instruction for a black-box LLM by optimizing the soft prompt of an open-source LLM using a Bayesian Optimization approach. To be more specific, within each Bayesian optimization iteration in InstructZero, a soft prompt is transformed into an instruction using the open-source LLM, and this instruction is subsequently fed into the black-box LLM. The output from the black-box LLM is then sent back to the Bayesian optimization process to generate the next soft prompt. For more details, please refer to Algorithm 1 in \citet{chen2023instructzero}.

\section{Implementation Details}
\label{app:imple_details}
As described in \Cref{sec:exp}, for a fair comparison, we allocate an equal budget to all methods. To elaborate further, APE and InstructZero search for the optimal prompt among 20 candidate instruction options, while in the case of MoP, the total number of candidate instructions across all experts sums up to 20. 

In the case of APE+Demos, APE+K-centroids, InstructZero+Demos, and InstructZero+K-centroids, each of them combines the best prompt found through APE or InstructZero with randomly selected demos or demos corresponding to centroids in the clustered embedding space. For these methods, the number of demos is set the same as in the case of MoP for a fair comparison.

Regarding hyperparameters, the $\alpha$ value in \Cref{eq:scaled_inertia} is set to the default value of 0.02 and remained the same across all experiments.

\section{Limitations}
\label{app:limitation}
To promote future exploration, we discuss two limitations of the proposed method.
First, the K-means-Auto algorithm used in the demo assignment does not guarantee the balance of the resulting clusters.
When a cluster receives demos that exceed the limit, we randomly discard them to meet the constraint.
This operation might be sub-optimal as it does not factor in their relative importance.
Future work might explore various data selection methods for trimming the cluster size.
Second, MoP uses existing instruction generation method (APE), but sometimes APE fails to generate sensible instructions in the first place. However, MoP can be applied to any instruction generation method, and if better instruction generation methods emerge in the future, we can also expect improved performance from MoP accordingly.
Lastly, the theory that motivates the use of clustering algorithm to assign demos - connecting ICL to kernel regression - cannot explain the demo order sensitivity in LLMs.
This suggests that future theoretical advancements could help motivate better demo assignment algorithms.
Finally, while the theory~\cite{han2023context} connecting ICL to kernel regression inspires the use of clustering algorithms for demo assignments, it could not explain the order sensitivity of demos in LLMs;
Further theoretical developments could help develop better demo assignment algorithms.

%% file: tables/rebuttal/more_baseline_bbh.tex
\begin{table*}[h!]
    \caption{\textbf{Comparison with OPRO and APE-Nearest-Neighbor on BIG-Bench-Hard.} We report the execution accuracy gain ($\Delta$) of MoP from the baseline described in \Cref{sec:exp:setting} on the BIG-Bench-Hard benchmark tasks. We run 3 experiments and provide both the mean and standard deviation values. Please note that due to the inherent randomness in the ChatGPT API, a performance gap of less than 1\% between the two methods can be considered a tie. The number of demos is set to $N_{\text{train}}/5$, where $N_{\text{train}}$ is the total number of training demos.}
    \begin{center}        
    \begin{tabular}{lccc}
        \toprule
        Task & APE-Nearest-Neighbor & OPRO & MoP \\
        \midrule
        \midrule
        Causal judgement & 59.93 $\pm$ 0.50 & 43.09 $\pm$ 5.85 & 60.99 $\pm$ 2.19 \\
        Disambiguation QA & 59.67 $\pm$ 0.47 & 48.00 $\pm$ 7.00 & 64.00 $\pm$ 2.16 \\
        Dyck languages & 14.00 $\pm$ 2.83 & 0.00 $\pm$ 0.00 & 17.33 $\pm$ 2.49 \\
        Movie Recommendation & 87.00 $\pm$ 2.45 & 70.50 $\pm$ 2.50 & 81.67 $\pm$ 2.36 \\
        Navigate & 47.33 $\pm$ 2.05 & 59.00 $\pm$ 4.00 & 54.00 $\pm$ 4.08 \\
        Object counting & 45.67 $\pm$ 1.70 & 60.00 $\pm$ 6.00 & 46.67 $\pm$ 2.05 \\
        Ruin names & 70.67 $\pm$ 2.62 & 70.00 $\pm$ 5.00 & 73.33 $\pm$ 1.89 \\
        Snarks & 56.55 $\pm$ 5.22 & 48.31 $\pm$ 3.37 & 55.81 $\pm$ 4.53 \\
        Sports understanding & 83.33 $\pm$ 0.94 & 23.50 $\pm$ 3.50 & 85.33 $\pm$ 2.62 \\
        Word sorting & 80.67 $\pm$ 1.25 & 63.00 $\pm$ 2.00 & 73.67 $\pm$ 1.89 \\
        \bottomrule

    \end{tabular}
    \end{center}
    \label{tbl:more_baseline_bbh}
\end{table*}

%% file: algos/mop.tex

\begin{algorithm}[H]
    \caption{\small Building MoP}
\begin{algorithmic}
    \STATE {\bfseries Input:} Training demos $\mathcal{D}^{\text{train}}={\{(x_i, y_i)\}}_{i=1}^{N^{\text{train}}}$, validation demos $\mathcal{D}^{\text{valid}}={\{(x_i, y_i)\}}_{i=1}^{N^{\text{valid}}}$, model $\mathcal{M}_{\phi}$, text encoder $\mathcal{E}_{\theta}(\cdot)$, task-specific scoring function $f(\cdot) \rightarrow \mathbb{R}$.
    
    \STATE \boxit{green}{5} \textit{$\triangleright$ Demo Assignment with clustering algorithm described in~\Cref{sec:method-demo_assign}}.
    \STATE {\bfseries Input:} $\alpha$ in \Cref{eq:scaled_inertia}, the minimum number of clusters $C_{\text{min}}$, the maximum number of clusters $C_{\text{max}}$
    \STATE Compute $\bm{e}_i^{\text{train}} = \mathcal{E}_{\theta}(x_i^{\text{train}})$ for $i=1, \dots, N_{\text{train}}$
    \STATE Select the best $C$ ($C^*$), which minimizes the scaled inertia score in~\Cref{eq:scaled_inertia}: 
    \STATE Clustering $\{\bm{e}_i^{\text{train}}\}_{i=1}^{N_{\text{train}}}$ into $C^*$ clusters.        
    \STATE {\bfseries Output:} Clustered demos $\{\mathcal{V}_1^{\text{train}}, \ldots, \mathcal{V}_{C^*}^{\text{train}}\}$ 

    \STATE \boxit{yellow}{7.1} \textit{$\triangleright$ Construct the region-based validation subset, $\mathcal{V}_c^{\text{valid}} \subset \mathcal{D}^{\text{valid}}$ using a clustering-based routing function.}
    \STATE {\bfseries Input:} $\{\mathcal{V}_1^{\text{train}}, \ldots, \mathcal{V}_{C^*}^{\text{train}}\}$ 
    \STATE $\mathcal{V}_c^{\text{valid}} \leftarrow \emptyset$ \quad for $c = 1, \ldots, C^*$ 
    \FOR{$i = 1$ {\bfseries to} $N_{\text{valid}}$}
        \STATE $c(x_{i}^{\text{valid}}) = \argmin_{c = 1, \ldots, C^*} K\left({\phi_{\theta}(x_{i}^{\text{valid}}), \bm{\mu}_{c} }\right).$ \hfill $\triangleright$ Routing function in \Cref{eq:routing} \hspace{2cm}\hspace*{\fill}
        \STATE $\mathcal{V}_{c}^{\text{valid}} \leftarrow \mathcal{V}_{c}^{\text{valid}} \cup \{({x_{i}^{\text{valid}}}, {y_{i}^{\text{valid}}})\}$ 
    \ENDFOR
    \STATE {\bfseries Output:} Clustered validation demos $\{\mathcal{V}_1^{\text{valid}}, \ldots, \mathcal{V}_{C^*}^{\text{valid}}\}$ 

    \STATE \boxit{blue}{9.3} \textit{$\triangleright$ Instruction Assignment with Region-based Joint Search described in~\Cref{sec:method-inst_assign}.}
    \STATE {\bfseries Input:} $\{\mathcal{V}_1^{\text{train}}, \ldots, \mathcal{V}_{C^*}^{\text{train}}\}$, $\{\mathcal{V}_1^{\text{valid}}, \ldots, \mathcal{V}_{C^*}^{\text{valid}}\}$ 
    \FOR{$c = 1$ {\bfseries to} $C^* $}
        \STATE Randomly sample a subset $\Tilde{\mathcal{D}}_{c}^{\text{train}} \sim \{ \mathcal{V}_1^{\text{train}}, \ldots, \mathcal{V}_{C^*}^{\text{train}} \} \setminus \{ \mathcal{V}_{c}^{\text{train}} \}$, where $|\Tilde{\mathcal{D}}_{c}^{\text{train}}| = r$.
        \STATE Generate candidate instructions $\{I_c^{j}\}_{j=1}^{m'}$ that complement the demos using a model $\mathcal{M}_\phi$ \\
        and a template format $T(\Tilde{\mathcal{D}}^{\text{train}})$ given $\Tilde{\mathcal{D}}^{\text{train}}$. 
        \STATE Evaluate the score on the region-based validation subset $\mathcal{V}_{c}^{\text{valid}}$ :
        \STATE $I_c^* = \argmax_{I_c^{j}} {\mathbb{E}_{(x, y) \sim \mathcal{V}_c^{\text{valid}}} {f([ I_{c}^{j}, {\mathcal{V}_{c}^{\text{train}}, x], y)}}}$
    \ENDFOR
    \STATE {\bfseries Output:} $\{I_{c}^*\}_{c=1}^{C^*}$
\STATE {\bfseries Output:} $\{P_c^*(x)\}_{c=1}^{C^*}$, where $P_c^*(x) = [I_c^*,\ \mathcal{V}_{c}^{\text{train}},\ x]$

\end{algorithmic}
\label{algo:mop}
\end{algorithm}

%% file: tables/rebuttal/search_cost.tex
\begin{table}[h]
\centering
\caption{\textbf{Comparison of search and inference costs for different algorithms}. The results are measured on the Instruction Induction Benchmark, using the same hardware (1$\times$ 48G A6000) and API model (\texttt{gpt-3.5-turbo-instruct}).}
    \begin{tabular}{llcc}
    \toprule
    \textbf{Method} & \textbf{Comment} & \textbf{Search cost (minute)} & \textbf{Inference cost (minute)} \\ \hline
    \midrule
    APE + demos & random demos & 0.22 & 0.013 / 100 queries \\ \hline
    IZ + demos & random demos & 8.87 & 0.012 / 100 queries \\ \hline
    MoP (10 experts) & w/o parallelization & 0.75 & 0.016 / 100 queries \\ \hline

    \bottomrule
    \end{tabular}
\label{tab:search_cost}
\end{table}


    

%% file: figures_tex/template.tex
\begin{figure}[t]
	\centering
    \includegraphics[width=0.9\linewidth]{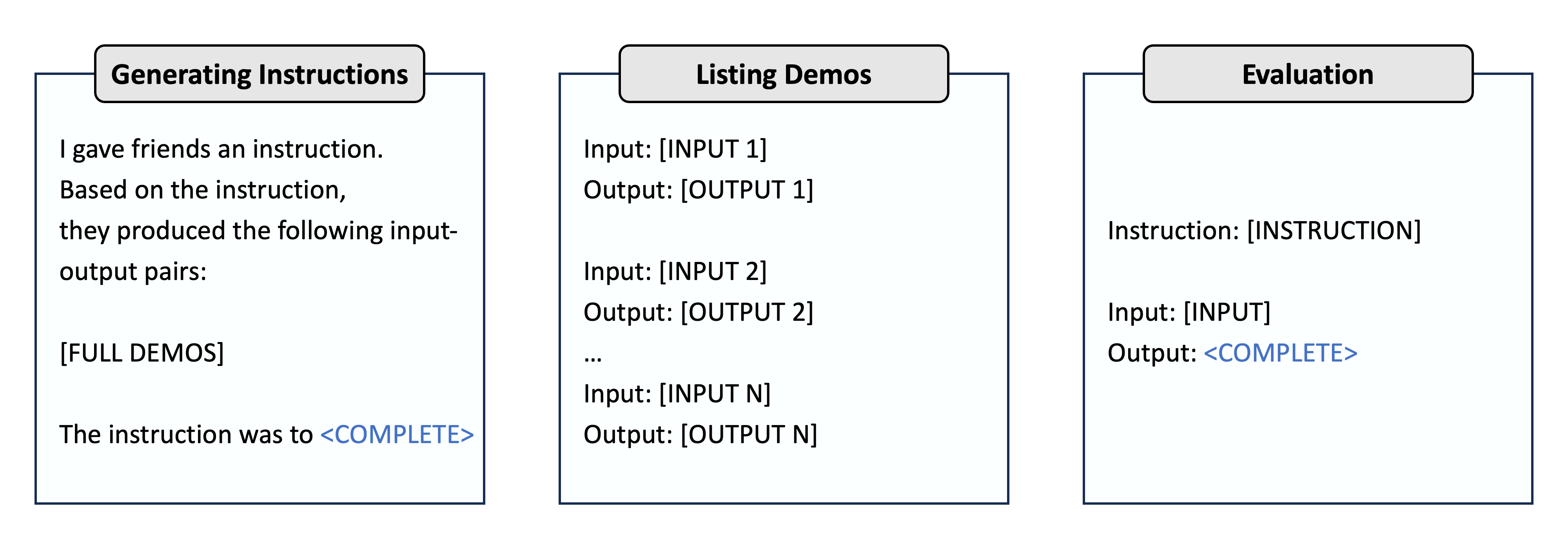}
	\caption{\textbf{The template used in our experiments.} }
 \label{fig:template}
\end{figure}

%% file: tables/rebuttal/ape_benchmark_description.tex
\begin{table}[t!]
\tiny
\centering
\caption{\textbf{Descriptions on Instruction Induction benchmark tasks.} Referring to~\citet{zhou2022large}, we provide task names, task summaries, and example demos within each task.}

\resizebox{0.92\textwidth}{!}{
    \begin{tabular}{p{2cm}p{5cm}p{5cm}}
    
    \toprule
    \textbf{Task} & \textbf{Task Summary} & \textbf{Demo} \\
    \midrule
    \midrule

    Auto categorization & Categorize items based on a common theme or characteristic. & Python, Cobol, and C $\rightarrow$ programming languages                       \\
    \midrule
    Rhymes               & Write a word that rhymes with the input word. & sing $\rightarrow$ ring                       \\
    \midrule
    Sentence similarity & Rate the semantic similarity of two input sentences on a scale of 0 - definitely not to 5 - perfectly. & Sentence 1: A man is smoking. Sentence 2: A man is skating. $\rightarrow$ 0 - definitely not  \\
    \midrule
    Sentiment            & Determine whether a movie review is positive or negative. & The film is small in scope, yet perfectly formed. $\rightarrow$ positive                     \\
    \midrule
    
    Word in context    &  Determine whether an input word has the same meaning in the two input sentences. &  Sentence 1: Approach a task. Sentence 2: To approach the city. Word: approach $\rightarrow$ not the same                         \\
    \midrule
    Larger animal       & Write the larger of the two given animals. & koala, snail $\rightarrow$ koala                         \\
    \midrule
    Informal to formal & Rephrase the sentence in formal language. & Please call once you get there $\rightarrow$ Please call upon your arrival.                          \\
    \midrule
    Orthography starts with & Extract the words starting with a given letter from the input sentence. & The man whose car I hit last week sued me. [m] $\rightarrow$ man, me                 \\
    \midrule
    Antonyms             & Write a word that means the opposite of the input word. & won $\rightarrow$ lost                              \\
    \midrule
    Second word letter & Extract the second letter of the input word. &  cat $\rightarrow$ a \\                      
    \midrule
    Common concept      & Find a common characteristic for the given objects. &  guitars, pendulums, neutrinos $\rightarrow$ involve oscillations                           \\
    \midrule
    Cause and effect   & Find which of the two given cause and effect sentences is the cause. & Sentence 1: The soda went flat. Sentence 2: The bottle was left open. $\rightarrow$ The bottle was left open.                    
    \\
    \midrule
    Translation EN-FR   & Translate the word into French. & time $\rightarrow$ temps                         \\
    \midrule
    Diff                 & Subtract the second number from the first. & 32 22 $\rightarrow$ 10                       \\
    \midrule
    First word letter  & Extract the first letter of the input word. &   cat $\rightarrow$ c                       \\
    \midrule
    Letters list        & Break the input word into letters, separated by spaces. & cat $\rightarrow$ c a t                                 \\
    \midrule
    Taxonomy animal     & Write all the animals that appear in the given list. & cat, helicopter, cook, whale, frog, lion $\rightarrow$ frog, cat, lion, whale                          \\
    \midrule
    Negation             & Negate the input sentence. & Time is finite $\rightarrow$ Time is not finite.                         \\
    \midrule
    Num to verbal      & Write the number in English words &  26 $\rightarrow$ twenty-six                          \\
    \midrule
    Active to passive  & Write the input sentence in passive form. & The artist introduced the scientist. $\rightarrow$ The scientist was introduced by the artist.                         \\
    \midrule
    Singular to plural & Convert the input word to its plural form. & cat $\rightarrow$ cats                        \\
    \midrule
    Sum                  &  Sum the two given numbers. & 22 10 $\rightarrow$ 32                            \\
    \midrule
    Synonyms             &  Write a word with a similar meaning to the input word. & alleged $\rightarrow$ supposed                         \\
    \midrule
    Translation EN-DE   & Translate the word into German. &  time $\rightarrow$ Zeit                         \\
    \midrule
    Translation EN-ES   &  Translate the word into Spanish. & time $\rightarrow$ hora                       \\
    \midrule
    Auto debugging      & Produce a specific result or output given the code. & import numpy as np $\backslash$n x = numpy.zeros(10) $\backslash$n $\rightarrow$ NameError: name 'numpy' is not defined.         \\
    \bottomrule
    \end{tabular}
}
\label{tab:ape_benchmark_description}
\end{table}


%% file: tables/superni/task_desc_code.tex
\newpage
\begin{table}[t!]
\tiny
\centering
\caption{\textbf{Descriptions on Super Natural Instructions benchmark code tasks.} Referring to \citet{superni}, we provide task names, task summaries, and example demos within each task.}

\resizebox{0.92\textwidth}{!}{
    \begin{tabular}{p{3cm}p{4cm}p{5cm}}
    
    \toprule
    \textbf{Task} & \textbf{Task Summary} & \textbf{Demo} \\
    \midrule
    \midrule

    Conala concat strings & Given a list of strings, concatenate them to form one string. & ['s', 'blew', 'g', 'and', 'u', 'as', 'C'] $\rightarrow$ sblewganduasC \\
    \midrule
    Conala normalize lists & Given a list of numbers, normalize the list such that the result adds to 1. & [-6.875, -64.545, -64.548] $\rightarrow$ [0.051 0.475 0.475] \\
    \midrule
    Conala calculate mean & Given a list of numbers, calculate the mean of the list. & [140.719, 220.491, 119.072] $\rightarrow$ 160.094 \\
    \midrule
    Conala max absolute value & Given a list of numbers, calculate the element with the largest absolute value. & [14.594 -85.985] $\rightarrow$ -85.985 \\
    \midrule
    Conala list index subtraction & Given a list of numbers, subtract each element by its index in the list. & [-14, 4] $\rightarrow$ [-15, 2] \\
    \midrule
    Conala remove duplicates & Given a list of numbers, remove all of the duplicates in the list. & [0, 0, 5, 7, 4, 3] $\rightarrow$ [5, 7, 4, 3] \\
    \midrule
    Conala list intersection & Given a two lists of numbers, find the intersection of the two lists. & [8, 10, 6, 2, 7] , [10, 4, 10, 10, 4] $\rightarrow$ [10] \\
    \midrule
    Splash question to sql & Generate an SQL statement from a question asking for certain data. & What are the names of all cartoons directed by Ben Jones? $\rightarrow$ SELECT Title FROM Cartoon WHERE Directed\_by = "Ben Jones" \\
    \midrule
    Logic2text sentence generation & Generate a natural language interpretation of the given logical operators. & most\_eq { all\_rows ; venue ; london } = true $\rightarrow$ for the venue records of all rows , most of them fuzzily match to london. \\
    \midrule
    Conala list index addition & Add lists together based on their index. & [[69, 8, -40], [63, -57, 65]] $\rightarrow$ [132, -49, 25] \\
    \midrule
    Conala sort dictionary & Sort a list of dictionaries based on a given key. & [{'first': 47, 'second': -34}, {'first': 11, 'second': 54}] $\rightarrow$ [{'first': 11, 'second': 54}, {'first': 47, 'second': -34}] \\
    \midrule
    Conala pair averages & Calculate the averages for each two consecutive elements. & [47, 62, 2, -13] $\rightarrow$ [54.5, 32.0, -5.5] \\
    \midrule
    Conala pair differences & Calculate the absolute difference for each two consecutive elements. & [-19, 40, 12, 95] $\rightarrow$ [59, 28, 83] \\
    \midrule
    English language answer relevance classification & Given a question and answer pair, detect whether the answer is acceptable or not. & Question: Is it more correct to say a computer program is, $\ldots$, Answer: I would say that neither, $\ldots$. $\rightarrow$ no \\
    \midrule
    Code x glue information retrieval & Given a code, calculate the number of for loops in the cpp program. & int ways(int n,int p), $\ldots$, $\rightarrow$ 1 \\
    
    \bottomrule
    \end{tabular}
}
\label{tab:superni_task_desc_code}
\end{table}

%% file: tables/superni/task_desc_math.tex
\newpage
\begin{table}[t!]
\tiny
\centering
\caption{\textbf{Descriptions on Super Natural Instructions benchmark mathematical tasks.} Referring to \citet{superni}, we provide task names, task summaries, and example demos within each task.}

\resizebox{0.92\textwidth}{!}{
    \begin{tabular}{p{3cm}p{4.1cm}p{4.9cm}}
    
    \toprule
    \textbf{Task} & \textbf{Task Summary} & \textbf{Demo} \\
    \midrule
    \midrule    

    Semeval 2019 task10 closed vocabulary mathematical answer generation & Answering multiple choices mathematical problem described with a closed-vocabulary. & If (frac{y}{y - 3} = frac{42}{39}), then what does y equal? (A) 39 (B) 41 (C) 42 (D) 45 (E) 81 $\rightarrow$ C \\
    \midrule
    Semeval 2019 task10 open vocabulary mathematical answer generation & Answering multiple choices mathematical problem described with an open vocabulary. & A new airplane can travel at speeds up to 4,680 miles per hour. How many miles can the airplane travel in 10 seconds? (A) 1.3 (B) 7.8 (C) 13 (D) 78 (E) 130 $\rightarrow$ C \\
    \midrule
    Ai2 arithmetic questions arithmetic & Given an arithmetic question, compute a solution. & Alyssa loves eating fruits. Alyssa paid \$12.05 for grapes, and \$9.85 for cherries. In total, how much money did Alyssa spend? $\rightarrow$ 21.9 \\ 
    \midrule
    Aqua multiple choice answering & Given a mathematical question, find the most suitable numerical answer. & Question: The sub-duplicate ratio of 16:64 is Option A: 4:3 Option B: 1:2 Option C: 1:3 Option D: 1:4 Option E: 2:4 $\rightarrow$ Option E \\ 
    \midrule
    Svamp subtraction question answering & Given a mathematical question involving subtraction, find the most suitable numerical answer. & Context: Baker sold 44 cakes. If he had made 48 cakes initially Question: How many cakes would baker still have? $\rightarrow$ 4 \\ 
    \midrule
    Mathdataset classification & Classify the type of a math word problem. & Solve 154 = -39*v - 41 for v. $\rightarrow$ algebra \\
    \midrule
    Mathdataset answer generation & Find the numerical answer for a math word problem. & Solve -38*s = -53*s - 90 for s. $\rightarrow$ -6 \\
    \midrule
    Asdiv addsub question answering & Given a mathematical question, find the most suitable numerical answer. & 46 apples were in the basket. 22 are red and the rest are green. how many apples are green? $\rightarrow$ 24 \\
    \midrule
    Asdiv multidiv question answering & Given a mathematical question, find the most suitable numerical answer. & each bag contains 23 pounds of oranges. how many pounds of oranges are in 45 bags? $\rightarrow$ 1035 \\
    \midrule
    Asdiv multiop question answering & Given a mathematical question, find the most suitable numerical answer. & a mirror store has 78 mirrors in stock. 8 mirrors are broken and 57 mirrors are sold. how many mirrors are left? $\rightarrow$ 13 \\
    \midrule
    Asdiv singleop question answering & Given a mathematical question, find the most suitable numerical answer. & nick saved \$68.50. if nick saved \$25.43 more than lee how much did lee save? $\rightarrow$ 43.07 \\
    \midrule
    Mawps addsub question answering & Given a mathematical question, find the most suitable numerical answer. & Mark has 13 trees in his backyard. If he plants 12 more, how many trees will he have? $\rightarrow$ 25 \\
    \midrule
    Mawps multidiv question answering & Given a mathematical question, find the most suitable numerical answer. & A cereal box holds 18 cups of cereal. Each serving is 2 cups. How many servings are in the whole box? $\rightarrow$ 9 \\
    \midrule
    Mawps multiop question answering & Given a mathematical question, find the most suitable numerical answer. & Paul had saved up 3 dollars. If he received another 7 dollars for his allowance, how many 5 dollar toys could he buy? $\rightarrow$ 2 \\
    \midrule
    Mawps singleop question answering & Given a mathematical question, find the most suitable numerical answer. & Joan has 9 blue balloons but lost 2 of them. How many blue balloons does Joan have now? $\rightarrow$ 7 \\
    \midrule
    Leetcode 420 strong password check & Check if the given password is strong & password = RtZGIgm7YeiPB66yVIoC $\rightarrow$ 0 \\
    \midrule
    Mathqa gain & Given a math problem on gain and options to choose from, find the correct option that answers the problem. & Problem: a 8\% stock yields 20\%. the market value of the stock is : Options: a) rs 48 , b) rs 45 , c) rs 40 , d) rs 50 , e) rs 55 $\rightarrow$ c \\
    \midrule
    Mathqa general & Given a general math problem and options to choose from, find the correct option that answers the problem. & Problem: what is the remainder of w = $3^19$ when divided by 10? Options: a) 0 , b) 1 , c) 5 , d) 7 , e) 9 $\rightarrow$ d \\
    \midrule
    Mathqa other & Given a math problem and options to choose from, find the correct option that answers the problem. & Problem: how many factors does $35^2$ have? Options: a) 2 , b) 8 , c) 24 , d) 25 , e) 26 $\rightarrow$ c \\
    \midrule
    Mathqa geometry & Given a problem on geometry and options to choose from, find the correct option that answers the problem. & Problem: the surface of a cube is 24 sq cm . find its volume? Options: a) 8 , b) 6 , c) 4 , d) 3 , e) 1 $\rightarrow$ a \\
    \midrule
    Mathqa probability & Given a problem on probability and options to choose from, find the correct option that answers the problem. & Problem: two coins are tossed. find the probability of at most 2 tails? Options: a) 1 / 2 , b) 1 / 4 , c) 1 / 3 , d) 1 , e) 3 / 4 $\rightarrow$ d \\
    \midrule
    Mathqa answer selection & Selecting answers to mathqa questions. & Problem: 1395 x 1395 Options: a. 1946025, b. 1981709, c. 18362619, d. 2031719, e. none of these $\rightarrow$ a \\
    \midrule
    Mathqa correct answer generation & Generate correct answers for math questions. & Problem: if 7 spiders make 7 webs in 7 days, then how many days are needed for 1 spider to make 1 web? $\rightarrow$ 7 \\
    
    \bottomrule
    \end{tabular}
}
\label{tab:superni_task_desc_math}
\end{table}

%% file: tables/bbh/task_desc.tex
\newpage
\begin{table}[t!]
\tiny
\centering
\caption{\textbf{Descriptions on BIG-Bench-Hard benchmark code tasks.} Referring to \citet{bbh}, we provide task names, task summaries, and example demos within each task.}

\resizebox{0.9\textwidth}{!}{
    \begin{tabular}{p{2.5cm}p{4.5cm}p{5cm}}
    
    \toprule
    \textbf{Task} & \textbf{Task Summary} & \textbf{Demo} \\
    \midrule
    \midrule    

    Causal judgement & Answer questions about causal attribution. & Frank T., had an ongoing dispute with his neighbor, $\ldots$. Did Frank T. intentionally shoot his neighbor in the body? Options: - Yes - No $\rightarrow$ Yes \\
    \midrule
    Disambiguation QA & Clarify the meaning of sentences with ambiguous pronouns. & Sentence: The scientist collaborated with the artist, and he shared a story. Options: (A) The scientist shared a story (B) The artist shared a story (C) Ambiguous $\rightarrow$ (C) \\
    \midrule
    Dyck languages & Correctly close a Dyck-n word. & Input: ( \{ \{ \} \} $\rightarrow$ ) \\
    \midrule
    Movie Recommendation & Recommend movies similar to the given list of movies. & Find a movie similar to Forrest Gump, The Silence of the Lambs, Seven, Fargo: Options: (A) Gandhi (B) Schindler's List (C) Dogfight (D) Repo Man $\rightarrow$ (B) \\
    \midrule
    Navigate & Given a series of navigation instructions, determine whether one would end up back at the starting point. & If you follow these instructions, do you return to the starting point? Take 5 steps. Take 4 steps. Take 3 steps. Options: - Yes - No $\rightarrow$ No\\
    \midrule
    Object Counting & Questions that involve enumerating objects of different types and asking the model to count them. & I have a piano, a flute, and four trombones. How many musical instruments do I have? $\rightarrow$ 6 \\
    \midrule
    Ruin names & Select the humorous edit that 'ruins' the input movie or musical artist name. & Which of the following is a humorous edit of this artist or movie name: 'bon iver'? Options: (A) bon liver (B) bion iver (C) ban iver (D) bon ivee $\rightarrow$ (A) \\
    \midrule
    Snarks & Determine which of two sentences is sarcastic. & Which statement is sarcastic? Options: (A) He's over six feet, so he must be tall (B) He's over six feet, so he must be wonderful $\rightarrow$ (B)\\
    \midrule
    Sports understanding & Determine whether an artificially constructed sentence relating to sports is plausible or implausible. & Is the following sentence plausible? "Mookie Betts skated behind the net." $\rightarrow$ no \\
    \midrule
    Word sorting & Sort a list of words. & List: thunderclap swab built poland $\rightarrow$ built poland swab thunderclap\\

    \bottomrule
    \end{tabular}
}
\label{tab:bbh_task_desc}
\end{table}

%% file: tables/ii/main_ii.tex
\begin{table*}[!ht]
\tiny
    \caption{\textbf{Results on Instruction Induction.} We report the execution accuracy gain ($\Delta$) of MoP from the baseline described in \Cref{sec:exp:setting} on the Instruction Induction benchmark tasks. We run 3 experiments and provide both the mean and standard deviation values. Please note that due to the inherent randomness in the ChatGPT API, a performance gap of less than 1\% between the two methods can be considered a tie. The number of demos is set to $N_{\text{train}}/10$, where $N_{\text{train}}$ is the total number of training demos.}
    \begin{center}
        
    \resizebox{\linewidth}{!}{
        \begin{tabular}{lS[table-number-alignment=left,table-format=3.2]S[table-number-alignment=left,table-format=3.2]S[table-number-alignment=left,table-format=3.2]S[table-number-alignment=left,table-format=3.2]S[table-number-alignment=left,table-format=3.2]S[table-number-alignment=left,table-format=3.2]}
            \toprule
            \multicolumn{1}{l}{\multirow{3}{*}{Task}} & \multicolumn{6}{c}{The execution accuracy gain ($\Delta$) of MoP from the following method (\%)} \\
                                                      & \multicolumn{1}{c}{APE} & \multicolumn{1}{c}{APE} & \multicolumn{1}{c}{APE} & \multicolumn{1}{c}{InstructZero} & \multicolumn{1}{c}{InstructZero} & \multicolumn{1}{c}{InstructZero} \\
                                                      & \multicolumn{1}{c}{\citep{zhou2022large}} & \multicolumn{1}{c}{+Demos} & \multicolumn{1}{c}{+K-centroids} & \multicolumn{1}{c}{\citep{chen2023instructzero}} & \multicolumn{1}{c}{+Demos} & \multicolumn{1}{c}{+K-centroids} \\
            
            \midrule
            \midrule

            Auto categorization & 35.33 \tiny$\pm$ 2.55 & 29.66 \tiny$\pm$ 2.48 & 26.00 \tiny$\pm$ 4.10 & 26.00 \tiny$\pm$ 2.81 & 22.66 \tiny$\pm$ 2.52 & 20.33 \tiny$\pm$ 2.92 \\
            Auto debugging & 29.17 \tiny$\pm$ 3.40 & 8.33 \tiny$\pm$ 3.40 & 0.00 \tiny$\pm$ 0.00 & 20.83 \tiny$\pm$ 3.40 & 12.50 \tiny$\pm$ 0.00 & 8.33 \tiny$\pm$ 3.40 \\
            Antonyms & 5.34 \tiny$\pm$ 3.08 & -0.66 \tiny$\pm$ 0.77 & -0.66 \tiny$\pm$ 0.61 & 3.00 \tiny$\pm$ 1.12 & 0.00 \tiny$\pm$ 0.77 & 0.00 \tiny$\pm$ 0.77 \\
            Cause and effect & 9.33 \tiny$\pm$ 9.30 & 6.66 \tiny$\pm$ 8.98 & 6.66 \tiny$\pm$ 9.17 & 9.33 \tiny$\pm$ 9.30 & 14.66 \tiny$\pm$ 7.93 & 17.33 \tiny$\pm$ 8.08 \\
            Common concept & 11.61 \tiny$\pm$ 2.95 & 6.05 \tiny$\pm$ 3.99 & -0.52 \tiny$\pm$ 3.70 & 6.68 \tiny$\pm$ 5.05 & 8.89 \tiny$\pm$ 4.51 & 0.39 \tiny$\pm$ 5.96 \\
            Informal to formal & -2.24 \tiny$\pm$ 2.55 & 2.61 \tiny$\pm$ 3.09 & 6.90 \tiny$\pm$ 2.97 & 13.89 \tiny$\pm$ 4.54 & 3.41 \tiny$\pm$ 3.77 & 13.56 \tiny$\pm$ 4.33 \\
            Taxonomy animal & 6.33 \tiny$\pm$ 4.15 & 20.00 \tiny$\pm$ 4.23 & 14.33 \tiny$\pm$ 5.22 & 19.66 \tiny$\pm$ 6.92 & 16.33 \tiny$\pm$ 7.29 & 10.33 \tiny$\pm$ 3.98 \\
            Negation & 4.33 \tiny$\pm$ 0.98 & 1.00 \tiny$\pm$ 0.90 & 1.66 \tiny$\pm$ 0.77 & 5.00 \tiny$\pm$ 2.29 & -0.67 \tiny$\pm$ 0.72 & 1.33 \tiny$\pm$ 1.09 \\
            Rhymes & -6.66 \tiny$\pm$ 15.07 & -0.66 \tiny$\pm$ 5.18 & -2.00 \tiny$\pm$ 8.12 & 17.67 \tiny$\pm$ 8.19 & 8.34 \tiny$\pm$ 6.34 & 3.34 \tiny$\pm$ 8.44 \\
            Sentence similarity & 32.00 \tiny$\pm$ 5.26 & 0.67 \tiny$\pm$ 4.03 & 5.00 \tiny$\pm$ 4.26 & 19.67 \tiny$\pm$ 8.89 & 1.34 \tiny$\pm$ 5.92 & 2.34 \tiny$\pm$ 6.44 \\
            Sentiment & 3.33 \tiny$\pm$ 0.55 & -0.33 \tiny$\pm$ 0.55 & 1.33 \tiny$\pm$ 0.55 & 5.33 \tiny$\pm$ 1.09 & 0.00 \tiny$\pm$ 0.67 & 0.67 \tiny$\pm$ 0.55 \\
            Orthography starts with & 4.00 \tiny$\pm$ 1.88 & -4.00 \tiny$\pm$ 1.88 & -0.67 \tiny$\pm$ 0.86 & 31.67 \tiny$\pm$ 13.74 & 4.33 \tiny$\pm$ 2.07 & 6.33 \tiny$\pm$ 2.51 \\
            Synonyms & 4.34 \tiny$\pm$ 1.54 & -2.33 \tiny$\pm$ 2.18 & -0.66 \tiny$\pm$ 2.03 & -12.00 \tiny$\pm$ 6.60 & 0.00 \tiny$\pm$ 1.46 & -3.00 \tiny$\pm$ 2.39 \\
            Translation EN-DE & 0.67 \tiny$\pm$ 0.72 & 1.67 \tiny$\pm$ 0.72 & -0.66 \tiny$\pm$ 0.90 & -1.00 \tiny$\pm$ 0.77 & 1.34 \tiny$\pm$ 0.77 & -0.66 \tiny$\pm$ 1.02 \\
            Translation EN-ES & 0.00 \tiny$\pm$ 1.02 & -0.67 \tiny$\pm$ 0.98 & -0.34 \tiny$\pm$ 1.02 & 1.66 \tiny$\pm$ 1.12 & 0.00 \tiny$\pm$ 1.02 & 0.00 \tiny$\pm$ 1.12 \\
            Translation EN-FR & 1.33 \tiny$\pm$ 1.09 & 0.33 \tiny$\pm$ 0.55 & -1.00 \tiny$\pm$ 0.67 & 3.33 \tiny$\pm$ 1.72 & 1.00 \tiny$\pm$ 0.94 & 0.67 \tiny$\pm$ 0.86 \\
            Word in context & 5.66 \tiny$\pm$ 2.52 & -3.67 \tiny$\pm$ 1.09 & 1.33 \tiny$\pm$ 1.09 & 10.66 \tiny$\pm$ 5.41 & -1.67 \tiny$\pm$ 1.19 & 1.66 \tiny$\pm$ 2.97 \\
            Diff & 0.00 \tiny$\pm$ 0.00 & 0.00 \tiny$\pm$ 0.00 & 0.00 \tiny$\pm$ 0.00 & 65.00 \tiny$\pm$ 26.54 &  0.00 \tiny$\pm$ 0.00 & 1.67 \tiny$\pm$ 1.36 \\
            First word letter & 0.00 \tiny$\pm$ 0.00 & 0.00 \tiny$\pm$ 0.00 & 0.00 \tiny$\pm$ 0.00 & 1.00 \tiny$\pm$ 0.47 & 0.00 \tiny$\pm$ 0.00 & 0.00 \tiny$\pm$ 0.00 \\
            Larger animal & -1.33 \tiny$\pm$ 0.72 & -1.00 \tiny$\pm$ 0.90 & 1.34 \tiny$\pm$ 1.22 & 22.67 \tiny$\pm$ 8.76 & 14.67 \tiny$\pm$ 6.72 & 14.67 \tiny$\pm$ 7.17 \\
            Letters list & 0.00 \tiny$\pm$ 0.00 & 0.00 \tiny$\pm$ 0.00 & 0.00 \tiny$\pm$ 0.00 & 0.00 \tiny$\pm$ 0.00 & 0.00 \tiny$\pm$ 0.00 & 0.00 \tiny$\pm$ 0.00 \\
            Num to verbal & 0.67 \tiny$\pm$ 0.27 & 0.00 \tiny$\pm$ 0.00 & 0.00 \tiny$\pm$ 0.00 & 1.00 \tiny$\pm$ 0.81 & 0.00 \tiny$\pm$ 0.00 & 0.00 \tiny$\pm$ 0.00 \\
            Active to passive & 0.00 \tiny$\pm$ 0.00 & 0.00 \tiny$\pm$ 0.00 & 0.00 \tiny$\pm$ 0.00 & 0.00 \tiny$\pm$ 0.00 & 0.00 \tiny$\pm$ 0.00 & 0.67 \tiny$\pm$ 0.54 \\
            Singular to plural & 2.34 \tiny$\pm$ 0.77 & 0.00 \tiny$\pm$ 0.38 & -0.33 \tiny$\pm$ 0.27 & -0.33 \tiny$\pm$ 0.27 & -0.33 \tiny$\pm$ 0.27 & -0.33 \tiny$\pm$ 0.27 \\
            Second word letter & -12.00 \tiny$\pm$ 9.80 & -12.00 \tiny$\pm$ 9.80 & -11.67 \tiny$\pm$ 9.80 & 25.33 \tiny$\pm$ 16.72 & 22.00 \tiny$\pm$ 16.94 & 21.00 \tiny$\pm$ 16.50 \\
            Sum & 0.00 \tiny$\pm$ 0.00 & 0.00 \tiny$\pm$ 0.00 & 0.00 \tiny$\pm$ 0.00 & 0.00 \tiny$\pm$ 0.00 & 0.00 \tiny$\pm$ 0.00 & 0.00 \tiny$\pm$ 0.00 \\

            \bottomrule
            \end{tabular}
        }
    \end{center}
    \label{tbl:main_ii}
\end{table*}

%% file: tables/superni/task_sel_criteria.tex
\begin{table*}[h!]
\tiny
    \caption{\textbf{The performance of APE on tasks related to code and mathematics in Super Natural Instructions.} We report the performance of APE~\citep{zhou2022large} on code and mathematics-related tasks, selected based on the domain information provided in the metadata of the Super Natural Instructions benchmark. We run 3 experiments and provide both the mean and standard deviation values. }
    \begin{center}
        
    \resizebox{0.6\linewidth}{!}{
        \begin{tabular}{lll}
            \toprule
            \multicolumn{2}{c}{\multirow{2}{*}{Task}} & \multicolumn{1}{c}{ROUGE-L (\%)} \\
                         &                             & \multicolumn{1}{c}{APE~\citep{zhou2022large}} \\
            
            \midrule
            \midrule

            \multirow{15}{*}{\rotatebox[origin=c]{90}{\tiny Code}} 

            & Conala concat strings & 88.84 \tiny$\pm$ 2.25 \\
            & Conala normalize lists & 45.37 \tiny$\pm$ 0.25 \\
            & Conala calculate mean & 23.00 \tiny$\pm$ 2.68 \\
            & Conala max absolute value & 23.00 \tiny$\pm$ 2.68 \\
            & Conala list index subtraction & 35.07 \tiny$\pm$ 9.55 \\
            & Conala remove duplicates & 72.22 \tiny$\pm$ 2.28 \\
            & Conala list intersection & 97.18 \tiny$\pm$ 0.35 \\
            & Splash question to sql & 60.54 \tiny$\pm$ 0.85 \\
            & Logic2text sentence generation & 41.26 \tiny$\pm$ 1.09 \\
            & Conala list index addition & 47.72 \tiny$\pm$ 10.39 \\
            & Conala sort dictionary & 99.84 \tiny$\pm$ 0.22 \\
            & Conala pair averages & 63.47 \tiny$\pm$ 11.05 \\
            & Conala pair differences & 18.86 \tiny$\pm$ 1.48 \\
            & English language answer relevance classification & 50.67 \tiny$\pm$ 2.49 \\
            & Code x glue information retreival & 20.52 \tiny$\pm$ 4.61 \\

            \hdashline        
            
            \multirow{25}{*}{\rotatebox[origin=c]{90}{\tiny Mathematics}}

            & Semeval 2019 task10 closed vocabulary & \multirow{2}{*}{18.60 \tiny$\pm$ 3.88} \\
            & mathematical answer generation & \\
            & Semeval 2019 task10 open vocabulary & \multirow{2}{*}{19.44 \tiny$\pm$ 3.61} \\
            & mathematical answer generation & \\
            & Ai2 arithmetic questions arithmetic & 51.81 \tiny$\pm$ 5.60 \\ 
            & Aqua multiple choice answering & 45.85 \tiny$\pm$ 0.59 \\ 
            & Svamp subtraction question answering & 67.79 \tiny$\pm$ 5.30 \\ 
            & Mathdataset classification & 46.11 \tiny$\pm$ 18.29 \\
            & Mathdataset answer generation & 24.11 \tiny$\pm$ 1.95 \\
            & Asdiv addsub question answering & 74.97 \tiny$\pm$ 11.30 \\
            & Asdiv multidiv question answering & 70.96 \tiny$\pm$ 10.94 \\
            & Asdiv multiop question answering & 73.01 \tiny$\pm$ 10.69 \\
            & Asdiv singleop question answering & 66.83 \tiny$\pm$ 13.95 \\
            & Mawps addsub question answering & 80.27 \tiny$\pm$ 4.64 \\
            & Mawps multidiv question answering & 52.17 \tiny$\pm$ 5.66 \\
            & Mawps multiop question answering & 59.69 \tiny$\pm$ 11.80 \\
            & Mawps singleop question answering & 77.30 \tiny$\pm$ 1.00 \\
            & Leetcode 420 strong password check & 6.66 \tiny$\pm$ 3.87 \\
            & Mathqa gain & 14.02 \tiny$\pm$ 2.04 \\
            & Mathqa general & 17.33 \tiny$\pm$ 1.30 \\
            & Mathqa other & 14.49 \tiny$\pm$ 2.43 \\
            & Mathqa geometry & 20.16 \tiny$\pm$ 1.01 \\
            & Mathqa probability & 12.08 \tiny$\pm$ 0.56 \\
            & Mathqa answer selection & 14.61 \tiny$\pm$ 1.32 \\
            & Mathqa correct answer generation & 27.15 \tiny$\pm$ 1.18 \\

            \bottomrule
            \end{tabular}
        }

    \end{center}
    \label{tbl:superni_task_sel_criteria}
\end{table*}

%% file: tables/superni/main_superni.tex
\begin{table*}[h!]
\tiny
    \caption{\textbf{Results for the Super-Natural Instructions.} We report the ROUGE-L score gain ($\Delta$) of MoP from the baseline described in \Cref{sec:exp:setting} on the Super Natural Instructions benchmark tasks. We run 3 experiments and provide both the mean and standard deviation values. Please note that due to the inherent randomness in the ChatGPT API, a performance gap of less than 1\% between the two methods can be considered a tie. The number of demos is set to $N_{\text{train}}/10$, where $N_{\text{train}}$ is the total number of training demos.}
    \begin{center}
        
    \resizebox{\linewidth}{!}{
        \begin{tabular}{llS[table-number-alignment=left,table-format=3.2,input-ignore]S[table-number-alignment=left,table-format=3.2,input-ignore]S[table-number-alignment=left,table-format=3.2,input-ignore]S[table-number-alignment=left,table-format=3.2,input-ignore]S[table-number-alignment=left,table-format=3.2,input-ignore]S[table-number-alignment=left,table-format=3.2,input-ignore]}
            \toprule
            \multicolumn{2}{c}{\multirow{2}{*}{Task}} & \multicolumn{6}{c}{The ROUGE-L score gain ($\Delta$) of MoP from the following method (\%)} \\
                                                      & & \multicolumn{1}{c}{APE} & \multicolumn{1}{c}{APE} & \multicolumn{1}{c}{APE} & \multicolumn{1}{c}{InstructZero} & \multicolumn{1}{c}{InstructZero} & \multicolumn{1}{c}{InstructZero} \\
                                                      & & \multicolumn{1}{c}{\citep{zhou2022large}} & \multicolumn{1}{c}{+Demos} & \multicolumn{1}{c}{+K-centroids} & \multicolumn{1}{c}{\citep{chen2023instructzero}} & \multicolumn{1}{c}{+Demos} & \multicolumn{1}{c}{+K-centroids} \\
            
            \midrule
            \midrule

            \multirow{7}{*}{\rotatebox[origin=c]{90}{\tiny Code}} 

            & Conala normalize lists & 1.98 \tiny$\pm$ 0.46 & 0.22 \tiny$\pm$ 0.54 & 1.33 \tiny$\pm$ 0.44 & 6.49 \tiny$\pm$ 2.09 & 0.60 \tiny$\pm$ 0.68 & 2.20 \tiny$\pm$ 0.76 \\
            & Conala calculate mean & 8.55 \tiny$\pm$ 2.38 & 6.05 \tiny$\pm$ 1.83 & 7.72 \tiny$\pm$ 1.85 & 7.38 \tiny$\pm$ 2.47 & 9.55 \tiny$\pm$ 2.58 & 9.65 \tiny$\pm$ 2.08 \\
            & Conala list index subtraction & 15.87 \tiny$\pm$ 7.76 & 14.53 \tiny$\pm$ 6.87 & 16.72 \tiny$\pm$ 6.31 & 35.20 \tiny$\pm$ 5.63 & 21.25 \tiny$\pm$ 6.44 & 21.44 \tiny$\pm$ 7.21 \\
            & Logic2text sentence generation & 55.22 \tiny$\pm$ 1.28 & 10.71 \tiny$\pm$ 1.14 & 2.68 \tiny$\pm$ 1.21 & 47.57 \tiny$\pm$ 9.28 & 8.71 \tiny$\pm$ 1.67 & 1.67 \tiny$\pm$ 1.13 \\
            & Conala list index addition & 4.47 \tiny$\pm$ 7.81 & -4.37 \tiny$\pm$ 6.11 & -0.01 \tiny$\pm$ 6.89 & 34.84 \tiny$\pm$ 5.13 & 24.45 \tiny$\pm$ 6.58 & 21.64 \tiny$\pm$ 6.76 \\
            & Conala pair differences & 58.91 \tiny$\pm$ 13.65 & 22.87 \tiny$\pm$ 18.05 & 17.12 \tiny$\pm$ 16.22 & 57.94 \tiny$\pm$ 15.04 & 44.32 \tiny$\pm$ 15.00 & 45.15 \tiny$\pm$ 13.78 \\
            & Code x glue information retreival & 19.48 \tiny$\pm$ 3.01 & -0.33 \tiny$\pm$ 2.64 & 4.67 \tiny$\pm$ 1.52 & 31.30 \tiny$\pm$ 2.63 & -0.67 \tiny$\pm$ 4.09 & 3.00 \tiny$\pm$ 1.63 \\

            \hdashline        
            
            \multirow{11}{*}{\rotatebox[origin=c]{90}{\tiny Mathematics}}

            & Semeval 2019 task10 closed voc. math. ans. gen. & 15.73 \tiny$\pm$ 2.53 & -2.67 \tiny$\pm$ 1.27 & -3.34 \tiny$\pm$ 1.54 & 10.37 \tiny$\pm$ 3.95 & -2.39 \tiny$\pm$ 1.89 & -0.27 \tiny$\pm$ 4.23 \\ 
            & Semeval 2019 task10 open voc. math. ans. gen. & 18.23 \tiny$\pm$ 2.21 & -0.66 \tiny$\pm$ 0.90 & -0.66 \tiny$\pm$ 2.34 & 12.64 \tiny$\pm$ 3.36 & -1.33 \tiny$\pm$ 0.86 & -1.00 \tiny$\pm$ 0.77 \\ 
            & Aqua multiple choice answering & 17.16 \tiny$\pm$ 0.71 & -6.16 \tiny$\pm$ 0.68 & -0.32 \tiny$\pm$ 0.72 & 19.27 \tiny$\pm$ 4.45 & -5.16 \tiny$\pm$ 0.72 & -1.49 \tiny$\pm$ 0.75 \\ 
            & Mathdataset classification & 49.22 \tiny$\pm$ 10.61 & 19.00 \tiny$\pm$ 2.34 & 7.00 \tiny$\pm$ 2.34 & 41.49 \tiny$\pm$ 2.69 & 20.66 \tiny$\pm$ 1.39 & 9.33 \tiny$\pm$ 2.28 \\
            & Mathdataset answer generation & 23.45 \tiny$\pm$ 2.08 & 4.89 \tiny$\pm$ 1.85 & 1.08 \tiny$\pm$ 2.30 & 21.44 \tiny$\pm$ 2.09 & 4.67 \tiny$\pm$ 2.09 & 0.58 \tiny$\pm$ 1.84 \\
            & Leetcode 420 strong password check & 20.67 \tiny$\pm$ 4.24 & 13.66 \tiny$\pm$ 3.67 & 2.33 \tiny$\pm$ 3.63 & 19.60 \tiny$\pm$ 5.24 & 8.66 \tiny$\pm$ 4.18 & 0.66 \tiny$\pm$ 5.35 \\
            & Mathqa gain & 19.31 \tiny$\pm$ 4.54 & 2.33 \tiny$\pm$ 4.41 & 4.00 \tiny$\pm$ 4.42 & 18.86 \tiny$\pm$ 4.50 & 3.33 \tiny$\pm$ 4.46 & 4.00 \tiny$\pm$ 4.39 \\
            & Mathqa general & 10.00 \tiny$\pm$ 1.24 & 4.66 \tiny$\pm$ 1.39 & 1.33 \tiny$\pm$ 2.28 & 12.73 \tiny$\pm$ 1.23 & 10.28 \tiny$\pm$ 4.57 & 7.30 \tiny$\pm$ 4.99 \\
            & Mathqa other & 21.18 \tiny$\pm$ 2.17 & 1.67 \tiny$\pm$ 1.72 & 5.67 \tiny$\pm$ 1.85 & 21.88 \tiny$\pm$ 2.16 & 3.34 \tiny$\pm$ 2.34 & 8.34 \tiny$\pm$ 1.81 \\
            & Mathqa geometry & 13.51 \tiny$\pm$ 0.80 & -5.33 \tiny$\pm$ 0.98 & 5.34 \tiny$\pm$ 0.61 & 12.58 \tiny$\pm$ 2.12 & -5.33 \tiny$\pm$ 0.72 & 5.34 \tiny$\pm$ 0.77 \\
            & Mathqa probability & 29.59 \tiny$\pm$ 0.63 & 10.00 \tiny$\pm$ 0.90 & 7.00 \tiny$\pm$ 0.90 & 24.61 \tiny$\pm$ 4.42 & 9.00 \tiny$\pm$ 0.90 & 7.67 \tiny$\pm$ 0.72 \\
            & Mathqa answer selection & 13.39 \tiny$\pm$ 1.61 & 0.33 \tiny$\pm$ 1.52 & -1.33 \tiny$\pm$ 2.96 & 13.64 \tiny$\pm$ 4.48 & 8.98 \tiny$\pm$ 7.48 & 4.72 \tiny$\pm$ 8.49 \\
            & Mathqa correct answer generation & 0.03 \tiny$\pm$ 2.50 & -3.90 \tiny$\pm$ 2.41 & 1.80 \tiny$\pm$ 2.41 & -3.52 \tiny$\pm$ 3.08 & -6.16 \tiny$\pm$ 2.42 & 0.49 \tiny$\pm$ 2.47 \\

            \bottomrule
            \end{tabular}
        }
    \end{center}
    \label{tbl:superni}
\end{table*}

%% file: tables/bbh/main_bbh.tex
\begin{table*}[h!]
\tiny
    \caption{\textbf{Results on BIG-Bench-Hard.} We report the execution accuracy gain ($\Delta$) of MoP from the baseline described in \Cref{sec:exp:setting} on the BIG-Bench-Hard benchmark tasks. We run 3 experiments and provide both the mean and standard deviation values. Please note that due to the inherent randomness in the ChatGPT API, a performance gap of less than 1\% between the two methods can be considered a tie. The number of demos is set to $N_{\text{train}}/5$, where $N_{\text{train}}$ is the total number of training demos.}
    \begin{center}
        
    \resizebox{\linewidth}{!}{
        \begin{tabular}{lS[table-number-alignment=left,table-format=3.2]S[table-number-alignment=left,table-format=3.2]S[table-number-alignment=left,table-format=3.2]S[table-number-alignment=left,table-format=3.2]S[table-number-alignment=left,table-format=3.2]S[table-number-alignment=left,table-format=3.2]}
            \toprule
            \multicolumn{1}{l}{\multirow{3}{*}{Task}} & \multicolumn{6}{c}{The execution accuracy gain ($\Delta$) of MoP from the following method (\%)} \\
                                                      & \multicolumn{1}{c}{APE} & \multicolumn{1}{c}{    APE    } & \multicolumn{1}{c}{    APE    } & \multicolumn{1}{c}{InstructZero} & \multicolumn{1}{c}{InstructZero} & \multicolumn{1}{c}{InstructZero} \\
                                                      & \multicolumn{1}{c}{\citep{zhou2022large}} & \multicolumn{1}{c}{+Demos} & \multicolumn{1}{c}{+K-centroids} & \multicolumn{1}{c}{\citep{chen2023instructzero}} & \multicolumn{1}{c}{+Demos} & \multicolumn{1}{c}{+K-centroids} \\
            
            \midrule
            \midrule

            Causal judgement & 6.38\tiny$\pm$ 1.64 & 1.42 \tiny$\pm$ 1.36 & 4.25 \tiny$\pm$ 1.92 & 0.35 \tiny$\pm$ 2.62 & 1.77 \tiny$\pm$ 1.30 &  1.77 \tiny$\pm$ 1.79 \\
            Disambiguation QA & 3.67 \tiny$\pm$ 2.95 & -1.67 \tiny$\pm$ 2.28 & -1.67 \tiny$\pm$ 1.72 & 7.00 \tiny$\pm$ 1.49 & -2.33 \tiny$\pm$ 1.28 &  -3.33 \tiny$\pm$ 2.07 \\
            Dyck languages & 9.33 \tiny$\pm$ 6.69 & 13.00 \tiny$\pm$ 1.61 & 5.33 \tiny$\pm$ 4.77 & 7.66 \tiny$\pm$ 6.49 & 6.66 \tiny$\pm$ 1.86 &  5.33 \tiny$\pm$ 5.54 \\
            Movie Recommendation & 11.67 \tiny$\pm$ 8.17 & 0.00 \tiny$\pm$ 1.98 & -2.33 \tiny$\pm$ 2.13 & 13.34 \tiny$\pm$ 5.63 & 3.67 \tiny$\pm$ 1.59 & 0.00 \tiny$\pm$ 2.52 \\
            Navigate & -5.00 \tiny$\pm$ 5.08 & 14.33 \tiny$\pm$ 2.46 & 4.00 \tiny$\pm$ 4.45 & 6.00 \tiny$\pm$ 2.87 & 14.00 \tiny$\pm$ 2.40 &  2.00 \tiny$\pm$ 5.93 \\
            Object counting & 0.34 \tiny$\pm$ 2.14 & 13.34 \tiny$\pm$ 1.21 & 3.67 \tiny$\pm$ 1.72 & 1.67 \tiny$\pm$ 5.38 & 12.00 \tiny$\pm$ 1.86 &  4.34 \tiny$\pm$ 1.86 \\
            Ruin names & 3.66 \tiny$\pm$ 1.87 & 6.00 \tiny$\pm$ 1.31 & 1.66 \tiny$\pm$ 1.75 & 5.66 \tiny$\pm$ 1.75 & 4.66 \tiny$\pm$ 2.25 &  3.66 \tiny$\pm$ 1.31 \\
            Snarks & -6.74 \tiny$\pm$ 3.99 & -0.37 \tiny$\pm$ 3.06 & -1.12 \tiny$\pm$ 2.84 & 2.25 \tiny$\pm$ 4.29 & 4.87 \tiny$\pm$ 3.92 & 4.12 \tiny$\pm$ 4.22 \\
            Sports understanding & 1.33 \tiny$\pm$ 2.27 & -1.67 \tiny$\pm$ 1.59 & 1.33 \tiny$\pm$ 2.80 & 5.00 \tiny$\pm$ 2.34 & -3.34 \tiny$\pm$ 1.61 & 0.00 \tiny$\pm$ 2.34 \\
            Word sorting & 10.67 \tiny$\pm$ 1.19 & -2.33 \tiny$\pm$ 1.09 & 1.00 \tiny$\pm$ 2.39 & 6.34 \tiny$\pm$ 5.16 & -4.00 \tiny$\pm$ 1.87 &  -2.00 \tiny$\pm$ 1.98 \\

            \bottomrule
            \end{tabular}
        }
    \end{center}
    \label{tbl:main_bbh}
\end{table*}

%% file: algos/ape.tex

\begin{algorithm}[H]
   \caption{APE~\citep{zhou2022large}}
   \label{algo:ape}
\begin{algorithmic}
   \STATE {\bfseries Input:} Training demos $\mathcal{D}^{\text{train}}={\{(x_i, y_i)\}}_{i=1}^{N^{\text{train}}}$, validation demos $\mathcal{D}^{\text{valid}}={\{(x_i, y_i)\}}_{i=1}^{N^{\text{valid}}}$, model $\mathcal{M}_\phi$, task-specific scoring function $f(\cdot) \rightarrow \mathbb{R}$.

   \STATE Randomly sample a subset $\Tilde{\mathcal{D}}^{\text{train}} \sim \mathcal{D}^{\text{train}}$, where $|\Tilde{\mathcal{D}}^{\text{train}}| = r$
   \STATE Generate candidate instructions $\{I^{j}\}_{j=1}^{m}$ using a model $\mathcal{M}_\phi$ and a template format $T(\Tilde{\mathcal{D}}^{\text{train}})$ given $\Tilde{\mathcal{D}}^{\text{train}}$ (\Cref{eq:gen_prompts}).
   \STATE Randomly sample a subset $\Tilde{\mathcal{D}}^{\text{valid}} \sim \mathcal{D}^{\text{valid}}$, where $|\Tilde{\mathcal{D}}_{\text{valid}}| = q$
   \FOR{$j=1$ {\bfseries to} $m$}
        \STATE Evaluate the instruction $I^{j}$ on the subset $\Tilde{\mathcal{D}}^{\text{valid}}$ and calculate the validation score; $\mathbb{E}_{(x, y) \sim \Tilde{\mathcal{D}}^{\text{valid}}}{f\left(\left[I^j,\ x\right], y\right)}$.
   \ENDFOR

   \STATE {\bfseries Output:} $P^*(x) = [I^*,\ x]$, where $I^* = \argmax_{I^j} {\mathbb{E}_{(x, y) \sim \Tilde{\mathcal{D}}^{\text{valid}}}{f\left(\left[I^j,\ x\right], y\right)}}.$  

\end{algorithmic}
\end{algorithm}